\documentclass[lettersize,journal]{IEEEtran}
\usepackage{amsfonts}
\usepackage{algorithmic}
\usepackage{algorithm}

\usepackage{pifont}
\usepackage{xcolor}
\usepackage{booktabs}
\usepackage{multirow}
\usepackage{makecell}
\usepackage{hhline}
\usepackage{amssymb}
\usepackage{arydshln}
\usepackage{array}
\usepackage[caption=false,font=normalsize,labelfont=sf,textfont=sf]{subfig}
\usepackage{textcomp}
\usepackage{stfloats}
\usepackage{url}
\usepackage{verbatim}
\usepackage{stfloats}
\usepackage{graphicx}
\usepackage{enumitem}
\usepackage{enumerate}

\usepackage{etoolbox}
\makeatletter
\patchcmd{\subsubsection}{\fi\ \@IEEEsectpunct}{\fi\ \@IEEEsectpunct}{}{}
\renewcommand{\@IEEEsectpunct}{\quad}
\makeatother
\usepackage{caption}

\usepackage{hyperref}

\usepackage{newtxtext}

\usepackage{csquotes}

\usepackage{mathtools}
\usepackage{amsmath}

\usepackage{orcidlink}

\begin{document}
\bstctlcite{IEEEexample:BSTcontrol}

\title{A Gift from the Integration of Discriminative and Diffusion-based Generative Learning: Boundary Refinement Remote Sensing Semantic Segmentation}

\author{Hao~Wang\,\raisebox{2pt}{\orcidlink{0000-0002-3964-479X}},
        Keyan~Hu\,\raisebox{2pt}{\orcidlink{0000-0003-0168-5606}},
        Xin~Guo\,\raisebox{2pt}{\orcidlink{0009-0009-6154-120X}},
        Haifeng~Li\,\raisebox{2pt}{\orcidlink{0000-0003-1173-6593}},~\IEEEmembership{Senior Member,~IEEE}
        and~Chao~Tao\,\raisebox{2pt}{\orcidlink{0000-0003-0071-310X}},~\IEEEmembership{Member,~IEEE}
\thanks{This article has been accepted for publication in the IEEE Transactions on Pattern Analysis and Machine Intelligence. This work was supported in part by the National Natural Science Foundation of China under Grant 42171376, Grant 42471419, and Grant 42501418 and in part by the High-Level Talent Introduction Project of Inner Mongolia University under Grant 10000-A25206020. The source code will be available at \url{https://github.com/KeyanHu-git/IDGBR}.}
\thanks{Hao Wang is with the School of Geosciences and Info-Physics, Central South University, Changsha 410083, China, and also with the College of Computer Science, Inner Mongolia University, Hohhot 010021, China.}
\thanks{Keyan Hu, Xin Guo, Haifeng Li, and Chao Tao are with the School of Geosciences and Info-Physics, Central South University, Changsha 410083, China.}
\thanks{Corresponding author: Chao Tao (email: kingtaochao@csu.edu.cn).}
}

\markboth{}{}

\maketitle

\begin{abstract}
Remote sensing semantic segmentation must address both what the ground objects are within an image and where they are located. Consequently, segmentation models must ensure not only the semantic correctness of large-scale patches (low-frequency information) but also the precise localization of boundaries between patches (high-frequency information related to boundary components). However, most existing approaches rely heavily on discriminative learning, which excels at capturing low-frequency features, while overlooking its inherent limitations in learning high-frequency features for semantic segmentation. Recent studies have revealed that diffusion generative models excel at generating high-frequency details. Our theoretical analysis confirms that the diffusion denoising process significantly enhances the model's ability to learn high-frequency features; however, we also observe that these models exhibit insufficient semantic inference for low-frequency features when guided solely by the original image. Therefore, we integrate the strengths of both discriminative and generative learning, proposing the Integration of Discriminative and diffusion-based Generative learning for Boundary Refinement (IDGBR) framework. The framework first generates a coarse segmentation map using a discriminative backbone model. This map and the original image are fed into a conditioning guidance network to jointly learn a guidance representation subsequently leveraged by an iterative denoising diffusion process refining the coarse segmentation. Extensive experiments across five remote sensing semantic segmentation datasets (binary and multi-class segmentation) confirm our framework’s capability of consistent boundary refinement for coarse results from diverse discriminative architectures.
\end{abstract}

\begin{IEEEkeywords}
Remote sensing semantic segmentation, boundary refinement, diffusion model, discriminative model, coarse segmentation refinement
\end{IEEEkeywords}

\section{Introduction}
\IEEEPARstart{S}{emantic} segmentation aims to classify each pixel in an image, thereby partitioning the image into multiple regions with specific semantic meanings. In remote sensing, semantic segmentation is also known as land use or land cover classification, representing a fundamental task in remote sensing image analysis. Semantic segmentation of remote sensing images provides crucial support for numerous remote sensing applications, including geographic information updates, environmental monitoring, and urban planning.
\begin{figure}[t!]
\centering
\includegraphics[width=3.5in]{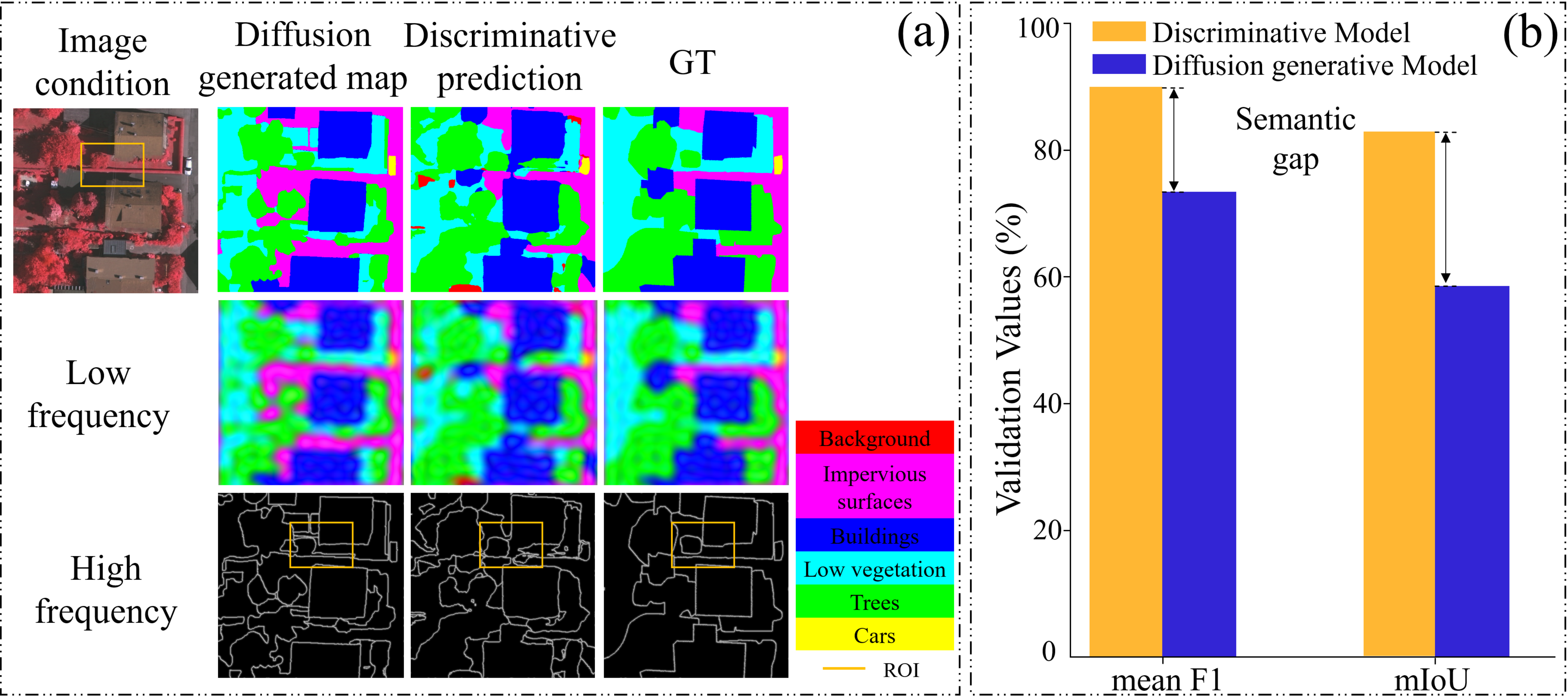}
\caption{A comparative evaluation between a diffusion model and a discriminative model for semantic segmentation of remote sensing imagery was conducted, with both models fully trained on the Potsdam dataset. The comparison is carried out from two perspectives: (a) visualization from a frequency-based viewpoint, and (b) overall semantic reasoning accuracy. The results reveal complementary characteristics between the two models: the diffusion model exhibits an inherent advantage in generating fine-grained boundaries (high-frequency information), while the discriminative model demonstrates stronger capability in semantic reasoning (low-frequency information).}
\label{fig1}
\end{figure}
In recent years, with the development of deep learning technology, semantic segmentation methods utilizing deep learning for remote sensing images have overcome the performance bottlenecks of traditional pixel-oriented or object-oriented segmentation approaches~\cite{heSwinTransformerEmbedding2022, hamaguchiEffectiveUseDilated2018}. Although these methods have significantly improved segmentation performance, they often fail to produce satisfactory boundary segmentation results~\cite{chenGeneralizedAsymmetricDualFront2021}. Most studies attribute the difficulty of boundary region segmentation to the fact that segmentation models rely on large-scale features to capture contextual information from spatial neighborhoods~\cite{marmanisClassificationEdgeImproving2018, kirillovPointRendImageSegmentation2020}. Compared to local features, large-scale features indeed provide richer semantic information. However, large-scale features are typically obtained from low-resolution feature maps, large receptive fields, long-range pixel relationships, or global pixel associations, thereby sacrificing local detail information. To address this issue, many studies have attempted to combine contextual semantic feature learning with boundary-sensitive local feature learning. Bertasius et al.~\cite{bertasiusConvolutionalRandomWalk2017} enhanced the model's boundary localization awareness by jointly optimizing pixel-wise affinity loss and semantic segmentation loss. Ding et al.~\cite{dingBoundaryAwareFeaturePropagation2019a} introduced an additional boundary category to be optimized along with semantic categories, and utilized a directed acyclic graph structure to use the learned boundaries to isolate contextual information propagation between different regions. Wang et al.~\cite{wangGeometricBoundaryGuided2023} designed a boundary-guided multi-scale feature fusion module to enhance boundary information preservation in multi-scale features. Considering the variability of boundary pixels between different categories, Wu et al.~\cite{wuConditionalBoundaryLoss2023} pulled each boundary pixel closer to its nearest local category center while pushing it away from surrounding categories.

However, existing studies primarily focus on enhancing intra-class similarity and inter-class separability in boundary regions under discriminative classification strategies, without adequately addressing the negative effects of discriminative learning on boundary segmentation. Boundaries refer to local transitions between adjacent different regions in the spatial domain. From a signal frequency perspective, boundaries are represented as high-frequency components in the frequency domain. Studies have shown that Convolutional Neural Networks (CNNs) can capture high-frequency components in image features~\cite{wangHighFrequencyComponentHelps2020}; however, discriminative classification tends to make CNNs prioritize learning low-frequency components while suppressing high-frequency ones~\cite{linInvestigatingExplainingFrequency2022}. This leads to insufficient feature learning for boundary pixels, thereby increasing the likelihood of their misclassification. From a probability theory perspective, discriminative classification directly learns the posterior distribution, disregarding the inherent characteristics of the data itself and only focusing on minimizing overall classification errors in the pixel space for given input data~\cite{ardizzoneTrainingNormalizingFlows2020}. Due to the small proportion of boundary pixels in the pixel space, errors in boundary pixel classification are often overlooked or even suppressed during the discriminative learning process~\cite{ayyoubzadehHighFrequencyDetail2021}.

Unlike discriminative learning, generative learning aims to produce high-fidelity samples by modeling the joint distribution of input data~\cite{ardizzoneTrainingNormalizingFlows2020}. Therefore, theoretically, generative learning needs to accurately model both low-frequency content structures and high-frequency boundary details in images, without neglecting or suppressing the learning of high-frequency components as discriminative learning does. However, for generative models such as VAEs and GANs, learning the distribution of high-frequency components remains challenging~\cite{schwarzFrequencyBiasGenerative2021}. In recent years, diffusion probabilistic models have achieved breakthrough progress in generative tasks by gradually adding noise to data and learning the reverse denoising process, thereby generating high-quality samples~\cite{rombachHighResolutionImageSynthesis2022a, songDenoisingDiffusionImplicit2021, hoDenoisingDiffusionProbabilistic2020}. In image generation tasks, studies have found that diffusion generative models enhance the learning of high-frequency components in images~\cite{siFreeUFreeLunch2024}. This paper also observes similar experimental phenomena in semantic segmentation tasks: by using remote sensing images as conditional information and applying diffusion noise training with Stable Diffusion~\cite{rombachHighResolutionImageSynthesis2022a} to segmentation label map encoded features, more realistic segmentation boundaries can be obtained. As shown in Fig.~\ref{fig1} (a), visualizing the high-frequency structural details extracted via Fourier transform reveals that diffusion models generate boundaries closer to the target geometric contours compared to discriminative models. However, from the perspective of overall segmentation metrics, there remains a significant gap in semantic reasoning accuracy between the segmentation results of diffusion generative models and discriminative models~\cite{yuRepresentationAlignmentGeneration2025}, as shown in Fig.~\ref{fig1} (b).

In summary, discriminative learning excels at capturing low-frequency large-scale semantic features but struggles to accurately learn high-frequency features; conversely, diffusion generative learning can effectively capture high-frequency features but lacks accuracy in inferring semantic features. The high-frequency information in images consists of multiple components, such as noise and object boundaries. In contrast, manually annotated semantic segmentation maps are generally free from image noise, and their high-frequency information is primarily concentrated in semantic boundary regions. Based on this analysis and observation, we propose the Integration of Discriminative and diffusion-based Generative learning for Boundary Refinement (IDGBR) framework, which learns the annotation patterns in boundary regions from manually labeled segmentation maps. The framework comprises a diffusion generative model, a discriminative model, and a conditional guidance network. First, to ensure semantic accuracy of the overall segmentation, IDGBR uses coarse segmentation results predicted by a discriminative model as semantic guidance conditions for diffusion generative learning. Although these coarse segmentation results have inaccurate boundary regions, they provide reliable semantic information for interior regions. Second, IDGBR captures joint representations from coarse segmentation maps and remote sensing images via a conditional guidance network. These representations are subsequently integrated into the diffusion model through residual connections to enhance semantic and boundary fidelity. Finally, we introduce a simple regularization technique that leverages guidance from a pre-trained representation model to explicitly align intermediate generative features. This technique is designed to mitigate initial training instability and to promote the intrinsic coherence of the semantic feature space.

Notably, current mainstream semantic segmentation evaluation metrics (e.g., mIoU and F1-score) struggle to effectively assess the precision and structural integrity of predictions in boundary regions. Their fundamental limitation lies in the reliance on pixel-level statistics, which neither account for spatial dependencies between pixels nor distinguish the varying importance of prediction errors between interior and boundary regions. Consequently, traditional metrics often fail to intuitively reflect substantive improvements in repairing boundary discontinuities, correcting boundary displacements, and sharpening blurred transitions. To overcome this evaluation limitation, we introduce the boundary-sensitive evaluation metric, Weighted F-measure (WFm)~\cite{margolinHowEvaluateForeground2014}. By incorporating a spatially sensitive weighting mechanism, this metric enables a more accurate and targeted quantification of performance gains in terms of boundary fidelity.

The main contributions of this paper are as follows:
\begin{itemize}
\item We demonstrate the effectiveness of boundary segmentation from the perspective of diffusion generative learning. Compared to discriminative models, diffusion generative models can better capture high-frequency boundary features, enabling them to generate more refined segmentation boundaries in remote sensing image semantic segmentation tasks.

\item We propose a remote sensing image semantic segmentation framework IDGBR for boundary refinement. IDGBR combines the advantages of discriminative learning and diffusion generative learning, learning joint guidance representations from coarse segmentation maps predicted by discriminative models and remote sensing images through a conditional guidance network, providing useful guidance for the diffusion generative model to learn semantic features and boundary features.

\item We introduce a regularization technique to explicitly align diffusion generative representations with joint guidance representations, enhancing the stability and semantic coherence of the feature space during initial training.

\item We evaluate the effectiveness of our method across different remote sensing scenarios through extensive experiments. Results on five semantic segmentation datasets demonstrate that the proposed IDGBR achieves significant improvements in boundary region segmentation.
\end{itemize}

\section{Related Work}
\subsection{General Semantic Segmentation}
Remote sensing semantic segmentation often faces challenges such as insufficient boundary refinement~\cite{DBLP:journals/pami/SuZWZLPL23, liSemanticSegmentationRemote2024, wuFSVLMVisionLanguageModel2025}, semantic ambiguity~\cite{heSwinTransformerEmbedding2022}, and missed segmentation of small-scale targets~\cite{kirillovPointRendImageSegmentation2020, huangMultiscaleSemanticSegmentation2025} due to large intra-class differences, low inter-class discrimination, and class imbalance in remote sensing features, as well as the large size, significant scale variations, and high scene complexity of remote sensing images. To address these issues, researchers have conducted a series of explorations.

CNNs represent the classical architecture in remote sensing semantic segmentation, with numerous research efforts building upon their strong local feature extraction capabilities to enhance the perception of multi-scale contextual information and boundary details~\cite{hamaguchiEffectiveUseDilated2018, liLSKNetFoundationLightweight2025}. Meanwhile, Transformers~\cite{vaswaniAttentionAllYou2017} have significantly enhanced the modeling of long-range dependencies in remote sensing imagery through their global self-attention mechanisms. For example, SETR~\cite{zhengRethinkingSemanticSegmentation2021} and Segmenter~\cite{strudelSegmenterTransformerSemantic2021a} reframe semantic segmentation as a sequence-to-sequence prediction task, pioneering a perspective that shifts from traditional convolutional stacking to global self-attention mechanisms. However, the quadratic complexity of self-attention mechanisms~\cite{vaswaniAttentionAllYou2017} renders them sensitive to training data scale and computational resources, prompting researchers to propose hybrid architectures that balance modeling capability with efficiency. Therefore, some studies, such as UNetFormer~\cite{wangUNetFormerUNetlikeTransformer2022} and ST-UNet~\cite{heSwinTransformerEmbedding2022}, have proposed hybrid architectures combining CNNs and Transformers to balance modeling capacity and computational efficiency. Nevertheless, Zhang et al.~\cite{zhangFsaNetFrequencySelfAttention2023} point out that both CNN-based and Transformer-based models tend to focus on low-frequency components in images, resulting in insufficient modeling of boundary information and fine-grained semantics.

\subsection{Boundary-Aware Semantic Segmentation}
\vspace{-0.2em}
To optimize model prediction accuracy in boundary regions, current research primarily focuses on designing boundary loss functions and improving model architectures. Regarding the design of boundary losses, typical approaches include geometry-aware guidance~\cite{wangGeometricBoundaryGuided2023}, distance-based inverse optimization~\cite{borseInverseFormLossFunction2021a}, and progressive alignment via differentiable vectors~\cite{wangActiveBoundaryLoss2022}. However, these methods are subject to two inherent limitations~\cite{wangActiveBoundaryLoss2022}. First, the optimization of boundary loss is local and posterior in nature. It typically operates only within the immediate vicinity of the model's current predictions, essentially acting as a local refinement of existing outputs. Second, boundary loss fundamentally conflicts with discriminative feature learning: while it aims to recover high-frequency details, discriminative models intrinsically favor smooth features. This contradiction leads to gradient conflicts between boundary and region losses, which often suppress boundary-related signals and result in training instability.

Regarding architectural improvements, methods can be generally categorized into two types: explicit boundary structure modeling and implicit boundary perception enhancement. The fundamental idea of explicit boundary structure modeling is to introduce a dedicated boundary processing branch within the model architecture to directly model and optimize boundary regions. These methods typically construct dual-stream network architectures, fusing conventional semantic features with boundary features through mechanisms such as gating mechanisms~\cite{takikawaGatedSCNNGatedShape2019}, feature propagation barrier constraints~\cite{dingBoundaryAwareFeaturePropagation2019a}, or pairwise semantic disentanglement~\cite{yinDisentangledNonlocalNeural2020}, thereby explicitly enhancing the model's perception of boundary structures. Conversely, the basic idea of implicit boundary perception enhancement is not to rely on parallel geometric structure branches, but to improve feature representation learning mechanisms, enabling the model to spontaneously learn more accurate boundary representations. These methods employ strategies such as local attention embedding~\cite{dingLANetLocalAttention2021}, graphics rendering concepts~\cite{kirillovPointRendImageSegmentation2020}, high-order affinity-based semantic decoupling~\cite{zhengHighOrderSemanticDecoupling2023}, task reformulation based on masked attention~\cite{chengMaskedAttentionMaskTransformer2022, raiMask2AnomalyMaskTransformer2024}, geometric prior generation~\cite{liSemanticSegmentationRemote2024}, or frequency-aware feature enhancement~\cite{zhangFsaNetFrequencySelfAttention2023, liFrequencyDecouplingNetwork2025} to implicitly enhance the learning capability of boundary regions in the feature space. However, the spectral bias and texture bias~\cite{geirhosImageNetTrainedCNNs2019} inherent in discriminative models are difficult to eliminate through architectural design alone. The fundamental reason lies in the optimization objective of discriminative learning, which seeks to minimize the overall error. As a result, low-frequency regions—which occupy most of the image area—dominate the gradient updates, forming an implicit inductive bias that favors low-frequency components. This bias drives the model to smooth sparse high-frequency boundaries, severely impairing the perception of global geometric topology. Consequently, architectural improvements based on discriminative models remain largely confined to ``neighborhood reasoning'' in terms of boundary enhancement performance.

\subsection{Diffusion-Based Generative Models}
In recent years, diffusion models have garnered widespread attention due to their powerful image generation capabilities. Some research has begun to incorporate diffusion models into segmentation tasks. For example, Baranchuk et al.~\cite{baranchukLabelEfficientSemanticSegmentation2022} discovered that intermediate activations in the reverse diffusion process of image generation can effectively capture pixel-level semantic representations for semantic segmentation. Zhao et al.~\cite{zhaoUnleashingTexttoImageDiffusion2023} have discovered implicitly aligned image-text representations in pretrained diffusion models and explored their applications in semantic segmentation tasks. However, from the perspective of overall segmentation metrics, there remains a significant gap in semantic reasoning accuracy between diffusion generative models and discriminative models~\cite{yuRepresentationAlignmentGeneration2025}. Addressing this issue, DDP~\cite{jiDDPDiffusionModel2023} replaced the traditional L2 loss in diffusion with cross-entropy loss and constructed a lightweight noise-to-prediction conditional diffusion-aware pipeline.

Similarly, in the remote sensing domain, diffusion models have demonstrated remarkable potential. In the context of remote sensing image generation, diffusion models have been successfully applied to high-fidelity remote sensing image synthesis under controllable conditions~\cite{khannaDiffusionSat2024}, continuous generation of global multi-resolution remote sensing imagery~\cite{ZhiMetaEarthGenerativeFoundation2025}, and cross-modal remote sensing image editing~\cite{liuText2Earth2025}. As for remote sensing image enhancement, Ref-Diff~\cite{dongBuildingBridges2024} and EDiffSR~\cite{xiaoEDiffSR2024} utilize ground object priors to explicitly guide the diffusion process, significantly improving the restoration quality of high-frequency details in super-resolution reconstruction. In summary, the successful application of diffusion models in the field of remote sensing stems from their powerful ability to learn high-frequency details, leading to superior performance over traditional methods in remote sensing image generation tasks.

We observe that discriminative learning excels at capturing low-frequency, large-scale semantic features but struggles to accurately learn high-frequency boundary features. Conversely, diffusion generative learning can effectively capture high-frequency boundary features but lacks accuracy in semantic feature inference. We combine the advantages of both discriminative learning and diffusion generative learning by implementing a conditional guidance network that learns joint guidance representations from coarse segmentation maps predicted by discriminative models and remote sensing images. This effectively guides the diffusion generative model to simultaneously focus on both global semantic structures and boundary details, significantly enhancing both the semantic integrity and boundary accuracy of segmentation results.

\section{Method}
In Section III-A, we first present the methodology for generative semantic segmentation using a latent diffusion model. In Section III-B, we analyze the denoising process of diffusion models from a frequency domain perspective, demonstrating their effectiveness in learning high-frequency components for semantic segmentation. In Section III-C, we introduce our IDGBR framework and detail how it learns joint guidance representations from coarse segmentation maps predicted by discriminative models and remote sensing images through a conditional guidance network, thereby providing effective guidance for the diffusion generative model. Finally, in Section III-D, we present a regularization technique that accelerates the learning process of generative semantic segmentation.

\subsection{Latent Diffusion Model for Generative Semantic Segmentation}
In recent years, diffusion models have achieved breakthrough progress in image generation tasks. The core idea of diffusion models is to gradually add noise to data and then learn how to reverse the process through denoising to generate new data samples. Unlike training diffusion models directly in high-dimensional pixel space, latent diffusion models (such as Stable Diffusion \cite{rombachHighResolutionImageSynthesis2022a}) use a pretrained encoder to compress images into smaller latent representations, and subsequently train diffusion models in the latent space, thereby reducing computational complexity. The latent diffusion model typically involves two data modeling processes: the diffusion process and the denoising process. The diffusion process gradually adds Gaussian noise to the latent representation of data, defined as:

\begin{equation}
q(z_t|z_0) = \mathcal{N}\bigl(z_t;\sqrt{\bar{\alpha}_t} z_0,(1-\bar{\alpha}_t)\mathbf{I}\bigr)
\label{eq.1}
\end{equation}

It progressively transforms the sample \( z_0 \) into a noisy sample \( z_t \), where \( t \in \{0,1,\dots,T\} \). The constant term \( \bar{\alpha}_t := \prod_{s=0}^t \alpha_s = \prod_{s=0}^t (1 - \beta_s) \), where \( \beta_s \) represents the noise schedule \cite{hoDenoisingDiffusionProbabilistic2020}. During training, the denoising process typically optimizes a Unet model \( \epsilon_\theta(z_t, c, t) \) by minimizing the mean squared error loss, which predicts the noise \( \varepsilon \) added to the noisy sample \( z_t \), under the guidance of condition \( c \):

\begin{equation}
\mathcal{L} = \mathbb{E}_{t,\varepsilon} \left[ \|\varepsilon - \epsilon_\theta(z_t, c, t)\|_2^2 \right]
\label{eq.2}
\end{equation}

During inference, random noise \( z_T \) is denoised through the denoising model \( \epsilon_\theta \) via a Markov chain under the guidance of condition \( c \), specifically \( z_T \to z_{T-1} \to \cdots \to z_0 \), which can be expressed as:

\begin{equation}
p_\theta(z_{0:T}|c) = p(z_T) \prod_{t=1}^T P_\theta(z_{t-1}|z_t,c)
\label{eq.3}
\end{equation}

However, latent diffusion models generally assume that sample data is continuous, making them unsuitable for processing discrete label maps. To address this, we introduce an auxiliary embedding network consisting of a shallow encoder $E_e$ and a decoder $D_e$. Specifically, $E_e$ projects discrete labels into a continuous embedding space to facilitate diffusion modeling, while $D_e$ maps the encoded features back to the category probability space. The optimization of the embedding network is independent of the diffusion model and does not require image data.

\begin{figure}[!t]
\centering
\includegraphics[width=3.4in]{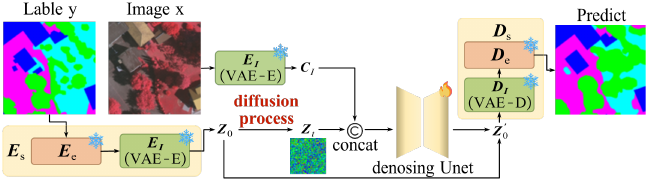}
\caption{Generative semantic segmentation pipeline using a latent diffusion model.}
\label{fig2}
\end{figure}

Given a remote sensing image \( x \in \mathbb{R}^{3 \times h \times w} \) and its corresponding segmentation label map \( y \in \mathbb{R}^{h \times w} \), the generative semantic segmentation process based on latent diffusion models is illustrated in Fig.~\ref{fig2}. Given the image encoder \( E_I \) and image decoder \( D_I \) from Stable Diffusion, and given the optimized shallow encoder \( E_e \) and shallow decoder \( D_e \) from the embedding network, we combine \( E_e \) with \( E_I \) to form encoder \( E_S \) for encoding discrete labels, and combine \( D_e \) with \( D_I \) to form decoder \( D_S \) for decoding label features. In our diffusion model training setup, the sample \( z_0 = E_S(y) \) and the conditioning information \( c_I = E_I(x) \). The diffusion process transforms the sample \( z_0 \) into a noisy sample \( z_t \) using Equation~\eqref{eq.1}. We concatenate \( z_t \) with \( c_I \) as input and train the Unet model \( \epsilon_\theta \) to predict the noise added to \( z_t \) using Equation~\eqref{eq.2}.

\subsection{Rethinking Diffusion Model from a Frequency Domain Perspective for Semantic Segmentation}
In the frequency domain, regions composed of similar neighboring pixels in an image typically manifest as low-frequency components, while edge regions with significant differences between adjacent pixels appear as high-frequency components. Similarly, when segmentation maps are transformed into the frequency domain via Fourier transform, the semantic content generally corresponds to low-frequency components, while boundaries between different semantic regions correspond to high-frequency components. Research has shown that diffusion generative models enhance the learning of high-frequency components in image generation tasks \cite{siFreeUFreeLunch2024, leeSpectrumTranslationRefinement2024}. Therefore, we explore whether diffusion models can enhance the refinement of object boundaries in semantic segmentation tasks. In this section, we first qualitatively evaluate the accuracy of diffusion-based generative semantic segmentation in inferring both high-frequency and low-frequency components. Subsequently, we theoretically analyze how the diffusion denoising process effectively enhances the model's ability to learn high-frequency components from input data.

\begin{figure*}[!t]
\centering
\includegraphics[width=1\textwidth]{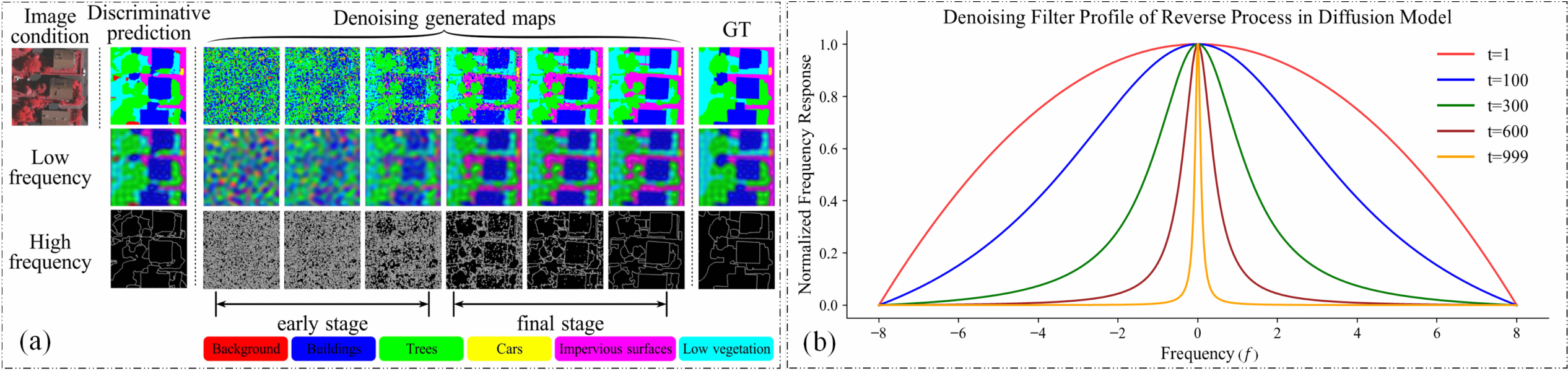}
\caption{Analysis of the effectiveness of diffusion-based generative learning for boundary segmentation. (a) Qualitative assessment of inference accuracy on high-frequency and low-frequency components. (b) Theoretical analysis of how the diffusion denoising process enhances high-frequency component learning. The horizontal axis represents relative spatial frequency ($f$), where smaller $|f|$ values correspond to low-frequency structures and larger $|f|$ values represent high-frequency details; the vertical axis is the normalized frequency response. The curves illustrate the dynamic evolution of the retention of different frequency components during the reverse denoising process.}
\label{fig3}
\end{figure*}

\subsubsection{Qualitative Evaluation of High/Low-Frequency Components in Diffusion-based Segmentation Maps}\leavevmode

We use remote sensing images as conditional information, employing the diffusion process to transform the encoded segmentation label samples \( z_0 \) into noisy samples \( z_t \), and train a Unet model to predict the noise added to \( z_t \). During denoising inference, we convert the segmentation maps generated from random noise \( z_T \) to the frequency domain via Fourier transform. For clearer observation, we divide the denoising process from \( z_T \) to \( z_0 \) into two sampling stages: the initial stage from \( z_T \) to \( z_{500} \) and the final stage from \( z_{500} \) to \( z_0 \), where \( T = 1000 \). Fig.~\ref{fig3} (a) illustrates two interesting phenomena we observed:

The high-frequency components in the generated segmentation maps continuously evolve throughout the entire denoising process, gradually transforming from initially rough and chaotic to refined, and ultimately approaching the geometric contours of the real targets. These high-frequency components reflect positions where semantic changes are most dramatic, such as boundary regions. When noise is introduced into segmentation maps, it typically manifests as random high frequencies that first contaminate the boundary regions, which are more sensitive to noise. To prevent high-frequency details from being weakened or lost during the generation process, the objective function of diffusion denoising enhances the model's ability to learn high-frequency components. Compared to segmentation maps predicted by discriminative models, those generated by diffusion models demonstrate significant advantages in boundary region refinement.

The low-frequency components in generated segmentation maps undergo significant changes during the initial stages of denoising but quickly stabilize, showing almost no change in the final stages of denoising. This indicates that the diffusion model's learning of low-frequency components primarily occurs in the initial sampling interval with high noise. While high-frequency components in segmentation maps represent positions of semantic mutations, the low-frequency components in different regions represent differentiated semantic content in remote sensing images. Therefore, the inference of high-frequency components and low-frequency components in segmentation maps corresponds to two distinct problems: Where do semantic mutations occur? What are the decision boundaries between different semantic categories?
Notably, the objective function of generative models does not directly learn decision boundaries between semantic categories, making diffusion generative models less accurate than discriminative models in low-frequency component inference.

\subsubsection{Why Diffusion Denoising Process Effectively Enhances High-Frequency Component Learning}\leavevmode

If we view the diffusion network as a linear filter, the optimal filter derived from the frequency response at each timestep is commonly known as the Wiener filter. Therefore, some studies have used Wiener filters to analyze the denoising process of diffusion models \cite{leeSpectrumTranslationRefinement2024, yangDiffusionProbabilisticModel2023}. In this subsection, we further explain the intrinsic reason why diffusion models effectively enhance high-frequency component learning based on the denoising process of the Wiener filtering method. Let $z_0$ denote a wide-sense stationary signal and $\varepsilon$ represent white noise with variance $\sigma^2 = 1$. For \( z_t = \sqrt{\bar{\alpha}_t} z_0 + \sqrt{1 - \bar{\alpha}_t} \varepsilon \), given a linear denoising Wiener filter \( h_t \), the objective function of the diffusion model at time \( t \) can be redefined as:

\begin{equation}
\mathcal{J}_t = \left\| \sqrt{\bar{\alpha}_t} z_0 - h_t * z_t \right\|^2
\label{eq.4}
\end{equation}

Minimizing \( \mathcal{J}_t \) in the frequency domain, we obtain the optimal solution of the Wiener filter \( h_t \) in the frequency domain, denoted as \( \mathcal{H}_t^*(f) \):

\begin{equation}
\mathcal{H}_t^*(f) = \frac{\bar{\alpha}_t |\mathcal{Z}_0(f)|^2}{\bar{\alpha}_t |\mathcal{Z}_0(f)|^2 + (1 - \bar{\alpha}_t) \cdot \sigma^2}
\label{eq.5}
\end{equation}
where \( |\mathcal{Z}_0(f)|^2 \) represents the power spectrum of \( z_0 \), and \( \mathcal{H}_t(f) \) represents the frequency response of \( h_t \). According to the power-law assumption commonly observed in image power spectra \cite{fieldRelationsStatisticsNatural1987}, we have \( \mathbb{E}[|\mathcal{Z}_0(f)|^2] \approx 1/f^2 \). Therefore:

\begin{equation}
\mathcal{H}_t^*(f) \approx \frac{\bar{\alpha}_t}{\bar{\alpha}_t + (1 - \bar{\alpha}_t) \cdot f^2}
\label{eq.6}
\end{equation}

We plot the normalized \( |\mathcal{H}_t^*(f)| \) curves in Fig.~\ref{fig3} (b). 
Detailed proof is provided in Appendix I.A. The denoising filter initially focuses only on generating low-frequency components during the early sampling process, and gradually enhances the generation of high-frequency components as \( t \) decreases. This indicates that to ensure signal reconstruction, diffusion models pay special attention to and learn high-frequency features in the signal, preventing important edge details from being weakened or lost during generation.

\subsection{The Integration Framework for Boundary Refinement Semantic Segmentation}

\begin{figure}[!t]
\centering
\includegraphics[width=3.4in]{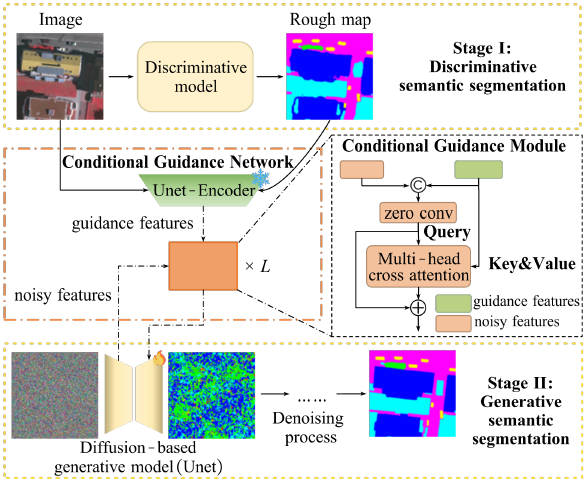}
\caption{Unified framework for remote sensing image segmentation merging discriminative models with diffusion generators.}
\label{fig4}
\end{figure}

In the previous subsection, we analyzed diffusion-based generative semantic segmentation from a frequency-domain perspective. Compared to discriminative learning, diffusion generative learning can effectively capture high-frequency boundary features but shows limitations in accurately inferring low-frequency semantic features. To fully leverage the boundary refinement capabilities of diffusion generative models while utilizing the large-scale semantic context from discriminative models, we propose IDGBR, a remote sensing semantic segmentation framework that integrates discriminative and generative models. As illustrated in Fig.~\ref{fig4}, the framework consists of a discriminative model, a diffusion generative model, and a conditional guidance network. Training consists of two stages. In the first stage, the discriminative model is trained to obtain coarse segmentation results, serving as conditional guidance for subsequent generative learning. In the second stage, under joint conditional guidance from remote sensing images and coarse segmentation maps, both the conditional guidance network and the diffusion generative model are trained to achieve boundary-refined semantic segmentation.

\subsubsection{Coarse Discriminative Semantic Segmentation as Conditional Guidance for Generative Learning}\leavevmode

Given a remote sensing image \( x \in \mathbb{R}^{3 \times h \times w} \) and its corresponding segmentation label map \( y \in \mathbb{R}^{h \times w} \), we first train a discriminative model \( d_\theta \) using cross-entropy loss. This discriminative model can be any conventional semantic segmentation model. We then use the trained discriminative model to generate coarse segmentation predictions:

\begin{equation}
y_r = d_\theta(x)
\label{eq.7}
\end{equation}

These coarse segmentation results serve as conditional guidance for generative learning, providing large-scale semantic context information to the generative semantic segmentation training process. In diffusion model training, the sample is defined as \( z_0 = E_S(y) \), and the conditional information as \( c_I = E_I(x) \) and \( c_r = E_S(y_r) \). The functions \( E_S(\cdot) \) and \( E_I(\cdot) \) denote encoding functions for segmentation maps and remote sensing images, respectively, aimed at transforming input data into low-dimensional latent encoding spaces to reduce computational complexity. The diffusion process transforms sample \( z_0 \) into a noisy sample \( z_t \). Under the joint conditional guidance of \( c_I \) and \( c_r \), we train a Unet model \( \epsilon_\theta \) to predict the noise contained in the noisy sample \( z_t \).

\subsubsection{Generative Boundary Refinement Semantic Segmentation under Conditional Guidance}\leavevmode

(a) Conditional Guidance Network:
To better extract joint guiding features from \( c_I \) and \( c_r \) and to facilitate generative semantic segmentation learning, we design a conditional guidance network. This network consists of a pseudo-siamese Unet encoder and several conditional guidance modules. By employing the pseudo-siamese network structure \cite{gaoCOMPOSECrossModalPseudoSiamese2020}, we can reuse parameters from the pretrained diffusion model's Unet encoder, and utilize long skip connections to pass guidance information to the training diffusion Unet decoder.

\begin{figure}[!t]
\centering
\includegraphics[width=3.4in]{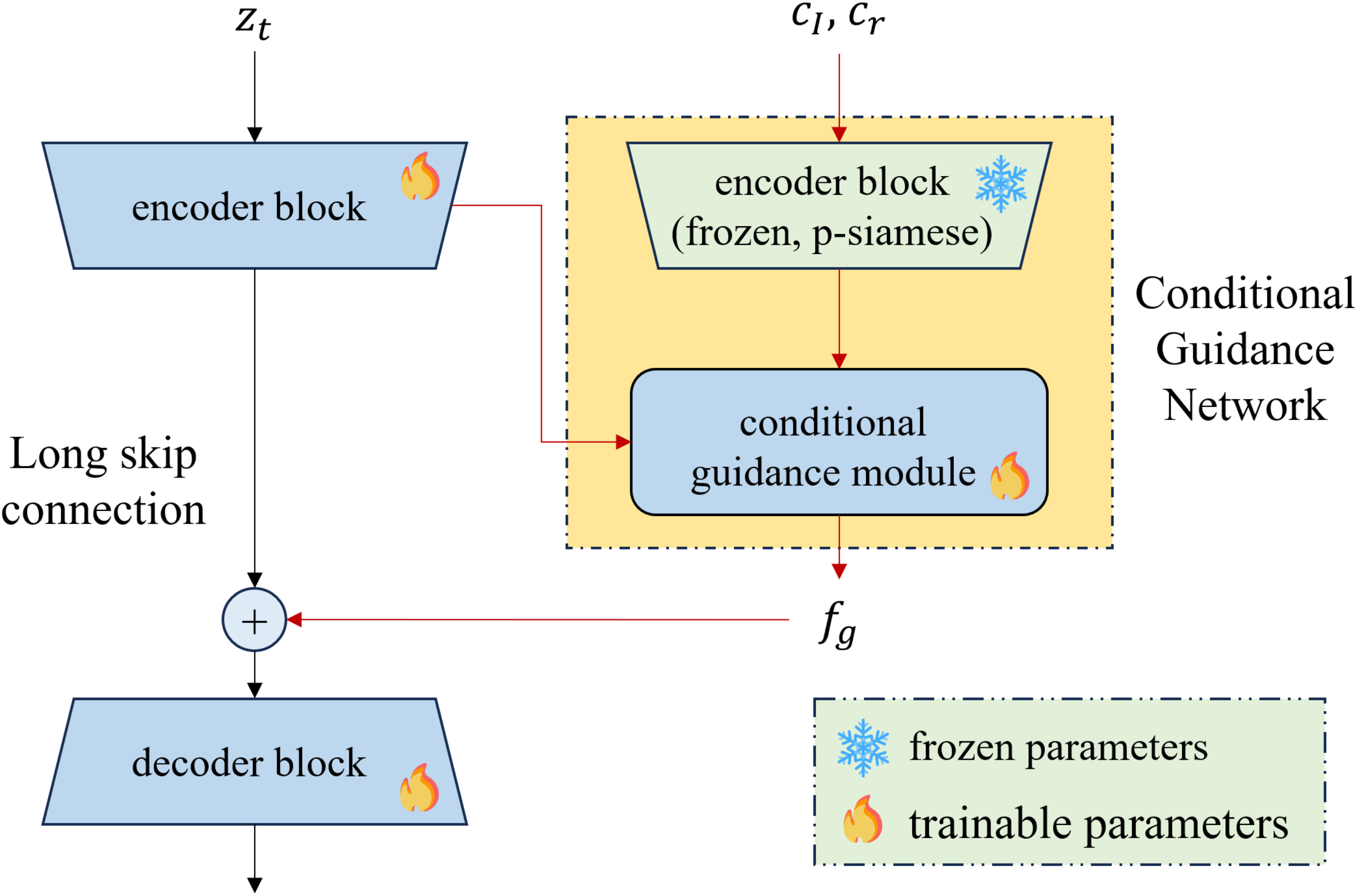}
\caption{Input and Output Data Flow of Conditional Guidance Network (Illustrated with an Encoding Module and its Corresponding Conditional Guidance Module)}
\label{fig5}
\end{figure}

We use Stable Diffusion v1.5 \cite{rombachHighResolutionImageSynthesis2022a} as the pretrained diffusion model. The Unet architecture in Stable Diffusion includes 12 encoding modules, 1 middle module, and 12 decoding modules connected via skip connections. The pseudo-siamese encoder in the conditional guidance network comprises 12 encoding modules and 1 middle module from Stable Diffusion's Unet. The outputs from these conditional guidance modules are passed via long skip connections to the 12 decoding modules and 1 middle module in the diffusion Unet. During diffusion Unet training, the pseudo-siamese encoder in the conditional guidance network reuses pretrained parameters and remains frozen; only the conditional guidance module parameters are trained. Fig.~\ref{fig5} illustrates the input and output data flow using one encoding module and its corresponding conditional guidance module in the conditional guidance network as an example.

(b) Conditional Guidance Modules:
Each conditional guidance module comprises a zero convolution layer and a cross-attention sub-module. The zero convolution layer merges noisy sample features with conditional guidance features by aggregating local information, capturing local correlations between these features. The cross-attention sub-module further explores global correlations within the fused features across channel dimensions, enhancing global interactions between noisy sample features and conditional guidance features.

Given a pretrained diffusion Unet model, we copy and freeze the encoder parameters from the pretrained Unet to serve as the pseudo-siamese encoder in the semantic guidance network, with parameters denoted as $\theta_c$. The parameters of the training diffusion Unet model are denoted as $\theta$. Let $\mathcal{F}(\cdot; \Theta)$ and $\mathcal{F}(\cdot; \Theta_c)$ represent the encoding network modules in the diffusion Unet and pseudo-siamese Unet, respectively. We denote the noisy sample features as $f_{z_t} = \mathcal{F}(z_t; \Theta)$ and the guidance features as $f_c = \text{concat}(\mathcal{F}(c_I; \Theta_c), \mathcal{F}(c_r; \Theta_c))$.

The diffusion Unet and pseudo-siamese Unet are connected through \( L \) conditional guidance modules (\( L = 13 \)), as shown in Fig.~\ref{fig4}. Each conditional interaction module consists of a \( 1 \times 1 \) zero convolution layer and a multi-head cross-attention layer. The zero convolution layer aims to capture local correlations between \( f_{z_t} \) and \( f_c \), initialized with zero weights, denoted as \( \text{ZConv}(\cdot) \). The output of the zero convolution layer is expressed as:

\begin{equation}
f' = \text{ZConv}(\text{Concat}(f_{z_t}, f_c))
\label{eq.8}
\end{equation}

After the convolution operation, we employ cross-attention to further capture global correlations. The multi-head cross-attention layer uses \( f' \) as the query \( Q \), and \( f_c \) as both key \( K \) and value \( V \), realizing:

\begin{equation}
\text{Attention}(Q, K, V) = \text{softmax}\left(\frac{QK^T}{\sqrt{d}}\right) \cdot V
\label{eq.9}
\end{equation}

where \( Q = W_Q \cdot \phi(f') \), \( K = W_K \cdot \phi(f_c) \), and \( V = W_V \cdot \phi(f_c) \). Here, \( W_Q \in \mathbb{R}^{d \times d_\epsilon} \), \( W_K \in \mathbb{R}^{d \times d_c} \), and \( W_V \in \mathbb{R}^{d \times d_c} \) are learnable projection matrices. \( \phi(f') \in \mathbb{R}^{N \times d_\epsilon} \) and \( \phi(f_c) \in \mathbb{R}^{N \times d_c} \) denote the flattening operation applied to \( f' \) and \( f_c \), respectively. The enhanced guidance features from the conditional interaction module are represented as:

\begin{equation}
f_g = f' + \text{Attention}(f', f_c, f_c)
\label{eq.10}
\end{equation}

Both \( f_g \) and \( f_{z_t} \) are transferred together through long skip connections to the decoder of the diffusion Unet.

\subsubsection{Regularization Strategy through Representation Alignment for Accelerated Training}\leavevmode

Although diffusion models excel at learning high-frequency patterns, their substantial training costs limit practical application. Inspired by REPA~\cite{yuRepresentationAlignmentGeneration2025}, we propose a regularization strategy based on representation alignment during early training stages. For semantic segmentation, this serves two key purposes: (1) it stabilizes early training by providing a robust semantic prior that guides optimization, leading to faster convergence; and (2) it enforces feature consistency across varying appearances and contexts, encouraging the model to reason based on semantic structure rather than superficial local textures. To this end, we select DINOv2~\cite{oquabDINOv2LearningRobust2024} as the semantic anchor, leveraging its strong general semantic representation—learned via self-distillation on large-scale unlabeled data—which focuses on scene geometry and object structure rather than low-level textures, thereby providing unbiased semantic guidance and avoiding task-specific bias. The implementation details are as follows:

(a) Visual Representation Extraction: Using pretrained visual encoder \( f \) and image \( x_c \), we extract spatial features: \( y_c = f(x_c) \in \mathbb{R}^{N \times D} \), where \( N \) represents the patch number and \( D \) is feature dimensionality.

(b) Diffusion Representation Alignment: At each timestep of the diffusion process, the final-layer hidden state from the Unet encoder is denoted by \( h_t = f_\theta(x_t) \). This hidden feature is transformed via a learnable MLP module \( h_\phi \) to align with \( y_c \).

\begin{equation}
\mathcal{L}_{\text{REPA}}(\theta, \phi) \coloneqq -\mathbb{E}_{x_c, \epsilon, t} \left[ \frac{1}{N} \sum_{n=1}^N \frac{y_c^{[n]} \cdot h_\phi(h_t^{[n]})}{\|y_c^{[n]}\| \|h_\phi(h_t^{[n]})\|} \right]
\label{eq.11}
\end{equation}
where n is a patch index and the cosine similarity is computed between corresponding patches.

\subsubsection{Training \& Inference}\leavevmode

\definecolor{mygreen}{HTML}{006400}
\definecolor{myblue}{HTML}{0000ff}
\begin{figure}[t]
    \centering
    \begin{minipage}{\columnwidth}
        \begin{algorithm}[H]
            \caption{IDGBR Training Process}\label{alg:idgbr}
            \begin{algorithmic}
                \STATE \textbf{{\color{myblue}Def}} train(imgs, labels, rough\_labels) 
                \STATE \hspace{0.5cm} {\color{mygreen}\# imgs: [b,3,h,w];} {\color{mygreen}labels \& }{\color{mygreen}rough\_labels: [b,1,h,w]}
                \STATE \hspace{0.5cm} {\color{mygreen}\# Step 1: Labels Encoding}
                \STATE \hspace{0.5cm} labels\_embedded = label\_embedding(labels)
                \STATE \hspace{0.5cm} rough\_embedded = label\_embedding(rough\_labels)
                \STATE \hspace{0.5cm} {\color{mygreen}\# Step 2: VAE Encoding}
                \STATE \hspace{0.5cm} img\_latents = vae\_encode(imgs)
                \STATE \hspace{0.5cm} label\_latents = vae\_encode(labels\_embedded)
                \STATE \hspace{0.5cm} rough\_latents = vae\_encode(rough\_embedded)
                \STATE \hspace{0.5cm} imgs\_repr = DINOv2\_encode(imgs)
                \STATE \hspace{0.5cm} {\color{mygreen}\# Step 3: Diffusion Process}
                \STATE \hspace{0.5cm} $T$ = 1000
                \STATE \hspace{0.5cm} $\epsilon$ = random\_like(label\_latents)
                \STATE \hspace{0.5cm} $t$ = uniform\_sample(0, $T$, size=(b,))
                \STATE \hspace{0.5cm} scaled\_t = clamp($T$ * (1 - (t / $T$)**3), 0, $T$-1)
                \STATE \hspace{0.5cm} $\epsilon$\_latents = DDIM(label\_latents, $\epsilon$, scaled\_t)
                \STATE \hspace{0.5cm} {\color{mygreen}\# Step 4: Model prediction}
                \STATE \hspace{0.5cm} preds\_$\epsilon$, latents\_repr = model($\epsilon$\_latents, 
                \STATE \hspace{2cm} img\_latents, rough\_latents, scaled\_t)
                \STATE \hspace{0.5cm} $\mathcal{L}_{\text{REPA}}$ = $-\mathbb{E}\!\bigl[\text{cosine\_similarity}(\text{latents\_repr},\,\text{imgs\_repr})\bigr]$
                \vspace{-0.4cm}
                \STATE \hspace{0.5cm} $\mathcal{L}_{\text{total}}$ = $\mathcal{L}_{\text{MSE}}$(preds\_$\epsilon$, $\epsilon$) + $\lambda$ $\cdot$ $\mathcal{L}_{\text{REPA}}$
                \STATE \hspace{0.5cm} \textbf{{\color{myblue}return}} $\mathcal{L}_{\text{total}}$
            \end{algorithmic}
        \end{algorithm}
    \end{minipage}
\end{figure}

(a) Training: During the training stage of the diffusion model, the objective is to train the denoising Unet model \(\epsilon_\theta\) to predict the noise \(\varepsilon\) added to the noisy sample \(z_t\), guided by conditional inputs \(c_I = E_I(x)\) and \(c_r = E_S(y_r)\). The detailed training algorithm is illustrated in Algorithm 1.

Label encoding: Diffusion models typically assume continuous input data, conflicting with discrete labels in semantic segmentation. Therefore, we design and pretrain a shallow label encoder \(E_e\) that maps discrete labels into a continuous differentiable latent space. Specifically, we construct a learnable embedding table for each class, transforming label map \(y \in \{0,1,\dots,K-1\}^{H \times W}\) into a continuous embedding feature \(E_0 \in \mathbb{R}^{C \times H \times W}\). This embedding is then normalized to the interval \([-1,1]\) by:

\begin{equation}
E = 2 \cdot \text{sigmoid}(E_0) - 1
\label{eq.12}
\end{equation}

The decoder \(D_e\), composed of two \(1 \times 1\) convolutional layers, maps the embedded features \(E\) back to the label space, producing class probability maps \(\hat{y} \in \mathbb{R}^{K \times H \times W}\). During training, we add Gaussian noise to the label embeddings to enhance the label decoder's ability to recover from embedding perturbations.

Noise schedule strategy: We employ the DDIM scheduler \cite{songDenoisingDiffusionImplicit2021}, which achieves deterministic sampling through reparameterization of the variance parameters in the diffusion process. We set the noise schedule parameter \(\beta_t\) to increase linearly from \(8.5 \times 10^{-4}\) to \(1.2 \times 10^{-2}\), covering \(T = 1000\) training timesteps. Furthermore, Mou et al. \cite{mouT2IAdapterLearningAdapters2024} discovered that enhancing conditional guidance during the early stages of the diffusion process significantly improves consistency between images and conditions, as well as the final generation quality. Inspired by this insight, to enhance the semantic perception capability of IDGBR during high-noise stages, we adopt a cubic timestep sampling strategy during training. The cubic timestep sampling formula is defined as follows:
\begin{equation}
t = \left(1 - \left(\frac{t}{T}\right)^3\right) \times T,\quad t \in U(0,T)
\label{13}
\end{equation}

Joint Training Objective: Under the Stable Diffusion v1.5 framework, we propose a dual-branch joint optimization strategy with a combined loss function consisting of the reconstruction error and representation alignment error. Specifically, this combined loss is defined as a weighted sum of the standard mean squared error loss \(\mathcal{L}_{\text{MSE}}\) and the representation alignment loss \(\mathcal{L}_{\text{REPA}}\):

\begin{equation}
\mathcal{L} = \mathcal{L}_{\text{MSE}} + \lambda \mathcal{L}_{\text{REPA}}
\label{14}
\end{equation}

Here, \(\mathcal{L}_{\text{MSE}}\) represents the standard denoising mean squared error loss, while \(\mathcal{L}_{\text{REPA}}\) is the representation alignment loss. During the first 200 training steps, we set \(\lambda = 0.5\) to incorporate semantic alignment; subsequently, \(\lambda\) is set to 0 to prevent continuous alignment signals from interfering with segmentation learning.

(b) Inference:  
During the inference stage of the diffusion model, we first sample an initial noise \( z_T \sim \mathcal{N}(0, \mathbf{I}) \) from a standard Gaussian distribution. Then, guided by conditions \( c_I = E_I(x) \) and \( c_r = E_S(y_r) \), we gradually reverse the diffusion process using the trained denoising network \( \epsilon_\theta \), recovering the noise-free semantic embedding \( z_0 \) through the DDIM sampler \cite{songDenoisingDiffusionImplicit2021}. Finally, we input \( z_0 \) into the label decoder to obtain the class prediction map \( \hat{y} \).

\definecolor{mygreen}{HTML}{006400}
\definecolor{myblue}{HTML}{0000ff}
\begin{figure}[t]
    \centering
    \begin{minipage}{\columnwidth}
        \begin{algorithm}[H]
            \caption{IDGBR Inferencing Process}\label{alg:idgbr-inference}
            \begin{algorithmic}
                \STATE \textbf{{\color{myblue}Def}} inference(imgs, rough\_label):
                \STATE \hspace{0.5cm} {\color{mygreen}\# Step 1: Labels Embedding}
                \STATE \hspace{0.5cm} embedded\_rough = label\_embedding(rough\_label)
                \STATE \hspace{0.5cm} {\color{mygreen}\# Step 2: VAE Encoding}
                \STATE \hspace{0.5cm} img\_latents = vae\_encode(imgs)
                \STATE \hspace{0.5cm} rough\_latents = vae\_encode(embedded\_rough)
                \STATE \hspace{0.5cm} {\color{mygreen}\# Step 3: Prepare timesteps}
                \STATE \hspace{0.5cm} scheduler.prediction\_type = $\epsilon$
                \STATE \hspace{0.5cm} timesteps = scheduler.get\_timesteps()
                \STATE \hspace{0.5cm} $\alpha$ = scheduler.alphas\_cumprod
                \STATE \hspace{0.5cm} {\color{mygreen}\# Step 4: Noise Initialization}
                \STATE \hspace{0.5cm} latents = random\_like(img\_latents)
                \STATE \hspace{0.5cm} {\color{mygreen}\# Step 5: Denoising Loop with DDIM update}
                \STATE \hspace{0.5cm} \textbf{{\color{myblue}for}} i \textbf{{\color{myblue}in}} range(len(timesteps) - 1):
                \STATE \hspace{1cm} $t$ = timesteps[i]
                \STATE \hspace{1cm} $t_{next}$ = timesteps[i + 1]
                \STATE \hspace{1cm} $\hat{\epsilon}$ = model(latents, img\_latents, rough\_latents, $t$)
                \STATE \hspace{1cm} $\bar{\alpha}_t = \alpha[t]$
                \STATE \hspace{1cm} $\bar{\alpha}_{t'} = \alpha[t_{next}]$
                \STATE \hspace{1cm} $\hat{x}_0 = (\text{latents} - \sqrt{1 - \bar{\alpha}_t} * \hat{\epsilon}) / \sqrt{\bar{\alpha}_t}$
                \STATE \hspace{1cm} latents = $\sqrt{\bar{\alpha}_{t'}} * \hat{x}_0 + \sqrt{1 - \bar{\alpha}_{t'}} * \hat{\epsilon}$
                \STATE \hspace{0.5cm} {\color{mygreen}\# Step 6: Decode to Segmentation Map}
                \STATE \hspace{0.5cm} decoded = vae\_decode(latents)
                \STATE \hspace{0.5cm} seg\_logits = label\_embedding\_decoder(decoded)
                \STATE \hspace{0.5cm} \textbf{{\color{myblue}return}} argmax(softmax(seg\_logits), dim=1)
            \end{algorithmic}
        \end{algorithm}
    \end{minipage}
\end{figure}

\section{Experiments}
The experiment establishes a comprehensive evaluation system to evaluate the effectiveness and generalizability of the proposed IDGBR semantic segmentation framework across different semantic segmentation tasks and discriminative base models. The evaluation covers representative binary and multi-class remote sensing segmentation tasks, with a focus on assessing IDGBR’s ability to refine high-frequency boundary information from coarse segmentation maps, enhance segmentation quality in complex scenes, and generalize effectively across diverse environments.

\subsection{Datasets}
The remote sensing datasets used in this paper can be divided into two types: binary and multi-class segmentation. The binary segmentation datasets include the CHN6-CUG road dataset \cite{zhuGlobalContextawareBatchindependent2021}, Fine-Grained Farmland Dataset (FGFD) \cite{liComprehensiveDeepLearningFramework2025}, and WHU Building change detection dataset \cite{jiFullyConvolutionalNetworks2019}. The multi-class segmentation datasets include the Potsdam \cite{ISPRS2DSemantic2022} and Vaihingen \cite{ISPRS2DSemantic2022a} urban feature classification datasets. Table~\ref{tab:table1} provides details about segmentation tasks for each dataset.

\begin{table}[h]
\centering
\caption{Remote sensing semantic segmentation datasets used in this paper\label{tab:table1}}
\label{tab:datasets}
\begin{tabular}{ll}
\toprule
\textbf{Dataset} & \textbf{Segmentation Tasks} \\
\midrule
CHN6-CUG & Road Segmentation \\
FGFD & Farmland Segmentation \\
WHU-Building & Building Segmentation \\
Potsdam & Landcover Segmentation (six categories)\\
Vaihingen & Landcover Segmentation (six categories)\\
\bottomrule
\end{tabular}
\end{table}

The CHN6-CUG road dataset aims to address the representation limitations of existing public datasets in complex road scenes. The dataset was collected from 6 typical urban areas and annotated by image interpretation experts to cover 12 road subclasses including railways, highways, urban roads, etc. It contains 4,511 images of 512 × 512 pixels at 1-meter resolution. In our experiments, all road subclasses were merged into one class, with non-road areas serving as the background class. The dataset was divided into training and test sets in a 4:1 ratio.

The FGFD addresses issues in agricultural remote sensing such as blurred farmland boundaries and severe terrain interference by constructing a refined dataset with sub-meter annotation accuracy. The dataset collects 0.3-meter resolution satellite images from China's seven major agricultural terrain regions (including 32.9\% plateau, 26.1\% plain, 8.6\% hills, etc.), covering an area of approximately 70 square kilometers. It employs a three-level quality control process of \enquote{annotation-cross-validation-expert sampling inspection} to generate 2,606 annotated samples of 512 × 512 pixels. In our experiments, the dataset was divided into training, validation, and test sets in an 8:1:1 ratio.

The WHU Building change detection dataset is constructed from aerial images downsampled to 0.3-meter resolution from two periods: 2012 (post-disaster) and 2016 (post-reconstruction). Geometric registration at 1.6-pixel accuracy was achieved using 30 ground control points. The original data covers 20.5 square kilometers, with annotated buildings increasing from 12,796 to 16,077, capturing the complete evolution of building clusters after the disaster. In our experiments, we cropped the images to 512 × 512 pixels, utilizing the post-disaster data for training and testing (7:3 ratio) and the post-reconstruction data exclusively for testing.

The Potsdam and Vaihingen datasets originate from the ISPRS Test Project on Urban Classification, 3D Building Reconstruction, and Semantic Labeling. Both datasets provide pixel-wise annotations for impervious surfaces, buildings, low vegetation, trees, cars, and background. The Potsdam dataset contains 38 aerial images with a ground sample distance (GSD) of 5 cm. Each image includes a true orthophoto (TOP) and a corresponding normalized digital surface model (nDSM). The orthophotos are available in three formats: infrared-red-green (IRRG), red-green-blue (RGB), and RGB-infrared composite, each at 6000 × 6000 resolution. The Vaihingen dataset contains 33 aerial images with a GSD of 9 cm, also including TOP and nDSM data. The TOP images in this dataset are provided in a near-infrared-red-green (NIR-RG) format with a resolution of 2494 × 2064. All images are cropped to 512 × 512 patches, yielding 3456 training and 2016 test images for Potsdam, and 562 training and 180 test images for Vaihingen.

\subsection{Experiment Settings}
\subsubsection{Implementation Details}\leavevmode

To provide a comprehensive evaluation of the generalization capability of IDGBR, we selected representative general segmentation methods covering three mainstream paradigms. By utilizing these models to generate the diverse coarse semantic priors required for boundary optimization, we verify the framework’s robustness across varying initial inputs: (1) Convolutional Neural Network methods: DeepLabV3+~\cite{chenEncoderDecoderAtrousSeparable2018} and the large-kernel CNN LSKNet~\cite{liLSKNetFoundationLightweight2025} optimized for remote sensing tasks; (2) Transformer methods: SegFormer~\cite{xieSegFormerSimpleEfficient2021}; (3) Self-supervised foundation models: using frozen DINOv2~\cite{oquabDINOv2LearningRobust2024} weights as the visual encoder.

For the baseline discriminative training, all models were implemented using open-source libraries with default hyperparameters. We employed standard data augmentation strategies: (1) random resizing with aspect ratio preserved between 0.5 and 2.0 (output size 512 × 512); (2) random cropping at size 512 ensuring no class exceeds 75\% coverage; (3) horizontal flipping with 50\% probability; (4) random adjustments to brightness, contrast, and saturation. All models were trained with a batch size of 8 for 20k iterations.

During the training of IDGBR, we employ a Unet network initialized with the pretrained weights of Stable Diffusion v1.5 as the backbone for the denoising process. Both images and labels are separately encoded using VAE encoders, and the resulting representations are concatenated to serve as conditional inputs. To avoid biases arising from model configuration, the training and inference procedures of IDGBR are kept consistent across different tasks and backbone models; for different numbers of classes, only the output projection dimension is modified during inference. Specifically, IDGBR model is trained for 80,000 iterations with a batch size of 4 and a fixed learning rate of 1e-5. Inference is performed over 25 steps, with a classifier-free guidance weight of 3. To accelerate model convergence, a representation alignment mechanism is introduced during early training. Concretely, we extract feature maps from the final layer of the Unet downsampling stage. These features are projected to semantic dimensions using a three-layer linear projection module, while a single linear expansion module handles spatial dimensional matching. The transformed features are then aligned with the semantic representations extracted by DINOv2, using cosine similarity as the alignment objective. This mechanism is only activated during the first 200 training steps to guide the model toward more efficient semantic learning.

\subsubsection{Evaluation Metrics}\leavevmode

In remote sensing semantic segmentation, the accuracy and integrity of boundaries are among the key factors ensuring the practical utility of mapping results. Margolin et al. \cite{margolinHowEvaluateForeground2014} pointed out that current mainstream evaluation metrics (such as IoU, mean F1, AP, and AUC) suffer from issues including ignoring the spatial dependencies between pixels and overlooking the varying importance of different prediction errors. Due to significant intra-class variability, blurred inter-class boundaries, challenges in foreground-background separation, and complex scenes inherent in remote sensing imagery, semantic segmentation tasks frequently exhibit insufficient boundary granularity and scattered patch fragments within segmented regions. These issues are prevalent and severely constrain the reliability of downstream tasks like object extraction and land cover statistics. This bottleneck stems from the lack of precise quantification of boundary structures in mainstream metrics, making it difficult to accurately capture model improvements in addressing boundary discontinuities, displacement, blurry transitions, or spurious patch corrections.

To overcome the limitations of the aforementioned evaluation assumptions, we incorporate the boundary-sensitive evaluation metric WF-measure \cite{margolinHowEvaluateForeground2014} (hereafter referred to as WFm). This metric extends the conventional $F_\beta$ measure to continuous domains and introduces a spatially sensitive weighting mechanism, enabling it to effectively assess the geometric fidelity of segmentation results. The calculation formula for WFm is as follows:

\begin{equation}
F_\beta^w = \frac{(1+\beta^2) \cdot \text{Precision}_w \cdot \text{Recall}_w}{\beta^2 \cdot \text{Precision}_w + \text{Recall}_w}
\label{eq.15}
\end{equation}

\begin{equation}
\text{Precision}_w = \frac{TP_w}{TP_w + FP_w}
\label{eq.16}
\end{equation}

\begin{equation}
\text{Recall}_w = \frac{TP_w}{TP_w + FN_w}
\label{eq.17}
\end{equation}

Among these, $\text{Precision}_w$ and $\text{Recall}_w$ represent weighted precision and weighted recall, respectively. The four weighted terms $TN_w$, $TP_w$, $FP_w$, and $FN_w$ are derived from traditional definitions via continuous value extension and incorporating of spatial dependency relationships and position-sensitive weight calculations. In our experiments, the boundary tolerance of WFm is set to 3 pixels (denoted as 3px) and the $\beta$ value is set to 1, providing a unified measure for evaluating the boundary quality of different models.

\subsection{Evaluation on Binary Segmentation Tasks}
In binary segmentation, we construct a validation framework using three representative binary remote sensing tasks: road segmentation, building extraction, and cropland delineation. These three tasks correspond to testing ground object features with linear extension, geometric regularity, and texture complexity, respectively, and effectively evaluate IDGBR's boundary reconstruction capability and target perception capabilities for objects with different morphological characteristics across various discriminative baseline models.

\subsubsection{Evaluation on Road Segmentation}\leavevmode

\begin{figure*}[!t]
\centering
\includegraphics[width=1\textwidth]{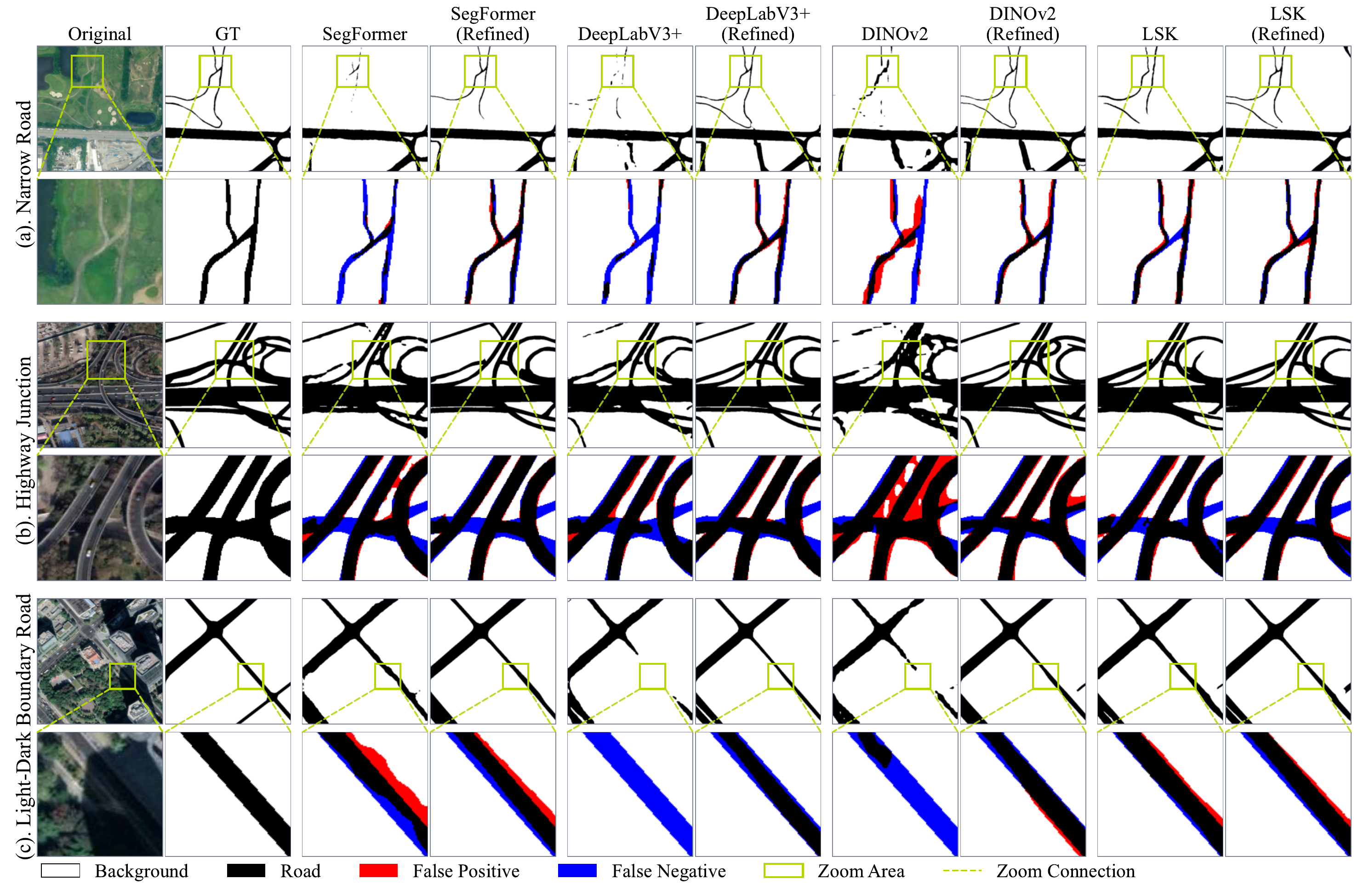}
\caption{Qualitative analysis on the CHN6-CUG road dataset, including DeepLabV3+, SegFormer, and DINOv2 methods and their results enhanced by IDGBR. Blue indicates false negatives (segmentation omissions), and red indicates false positives (segmentation commissions).}
\label{fig6}
\end{figure*}

The complex topological structures and multi-scale characteristics of roads in the CHN6-CUG dataset serve as a rigorous benchmark for assessing the boundary refinement and connectivity restoration for linear features. Comparative results are presented as follows:

\begin{table}[htbp]
\centering
\caption{Quantitative comparison (\%) of our proposed method against various general segmentation methods on the CHN6-CUG road dataset.}
\label{tab:performance_road}
\begingroup
\footnotesize
\setlength{\tabcolsep}{3pt}
\renewcommand{\arraystretch}{1}
\setlength{\dashlinedash}{2pt}
\setlength{\dashlinegap}{1.5pt}
\resizebox{\columnwidth}{!}{
\begin{tabular}{l|l|cc|c @{\hspace{10pt}} l}
\noalign{\hrule height 1pt}
\multirow{2}{*}{Method}
& \multirow{2}{*}{\begin{tabular}[c]{@{}c@{}}DiscModel\\[-0.3ex]Backbone\end{tabular}}
& \multicolumn{2}{c|}{Road}
& \multirow{2}{*}{Acc}
& \multirow{2}{*}{\makecell[l]{WFm\\[-0.5ex](3px B)}} \\
\cline{3-4}
& & IoU & F1-score & & \\ \hline
\noalign{\vskip 0.7pt}
DeepLabV3+ & ResNet-50 & 62.59 & 76.99 & 97.48 & 34.44 \\
DeepLabV3+$\rightarrow$ours & ResNet-50 & 62.94 & 77.25 & 97.34 & 36.49$_{\text{+2.05}}$ \\
SegFormer & MIT-B5 & 64.13 & 78.14 & 97.50 & 36.00 \\
SegFormer$\rightarrow$ours & MIT-B5 & 64.54 & 78.45 & 97.50 & \textbf{37.64}$_{\text{+1.64}}$ \\
DINOv2 & ViT-B/14 & 57.37 & 72.91 & 96.96 & 27.93 \\
DINOv2$\rightarrow$ours & ViT-B/14 & 61.07 & 75.83 & 97.17 & 34.65$_{\text{+6.72}}$ \\
LSK & LSK-s & 62.46 & 76.89 & 97.44 & 35.90 \\
LSK$\rightarrow$ours & LSK-s & 63.30 & 77.53 & 97.47 & 37.31$_{\text{+1.41}}$ \\
\noalign{\hrule height 1pt}
\end{tabular}}
\endgroup
\end{table}

As shown in Table~\ref{tab:performance_road}, analysis based on the WFm boundary metric demonstrates that optimizing boundaries with the proposed IDGBR framework leads to significant improvements in boundary scores for models such as DeepLabV3+, SegFormer, DINOv2, and LSK, indicating strong cross-model generalization capability. In terms of the IoU metric, all baseline models achieved consistent improvements after boundary optimization, with DINOv2 achieving the most notable gains.

Discriminative models commonly exhibit three typical deficiencies when dealing with complex road scenarios: (1) Insufficient precision in high-frequency edge segmentation, (2) Incomplete delineation of topological structures in complex road networks, and (3) Inadequate preservation of road network structural continuity. As shown in Fig.~\ref{fig6}, we perform a comparative analysis of different segmentation models across three representative scenarios. In narrow road scenarios shown in Fig.~\ref{fig6} (a), baseline discriminative models often suffer from significant omission errors, demonstrating their substantially limited ability to perceive low-contrast elongated structures. In contrast, our optimized model can effectively enhance the perception of such low-visibility features based on coarse segmentation results, achieving significant improvements in segmentation completeness. Fig.~\ref{fig6} (b) illustrates a complex highway scene with intersections and overpasses. Discriminative models struggle to maintain road integrity and connectivity when handling overlapping structures and complex topological structures, leading to incomplete geometric features of the extracted roads. Our approach refines road edge details and supplements topological relationships based on the initial coarse predictions. Fig.~\ref{fig6} (c) illustrates road scenarios under high-contrast illumination conditions. Under strong illumination gradients, drastic high-frequency variations tend to mislead discriminative models into overfitting to artifact features at light-shadow boundaries, resulting in structural omissions and topological misinterpretations in shaded areas. Through iterative boundary optimization, our method significantly improves the edge localization precision of baseline models in unevenly illuminated regions with complex shadows.

\subsubsection{Evaluation on Cultivated Land Segmentation}\leavevmode

\begin{table}[htbp]
  \centering
  \caption{Quantitative comparison (\%) of our proposed method against various general segmentation methods on the FGFD.}
  \label{tab:performance_fgfd}
  \begingroup
    \footnotesize
    \setlength{\tabcolsep}{3pt}
    \renewcommand{\arraystretch}{1}
    \setlength{\dashlinedash}{2pt}
    \setlength{\dashlinegap}{1.5pt}
    \resizebox{\columnwidth}{!}{
      \begin{tabular}{l|l|cc|c @{\hspace{10pt}} l}
        \noalign{\hrule height 1pt}
        \multirow{2}{*}{Method}
          & \multirow{2}{*}{\begin{tabular}[c]{@{}c@{}}DiscModel\\[-0.3ex]Backbone\end{tabular}}
          & \multicolumn{2}{c|}{Farmland}
          & \multirow{2}{*}{Acc}
          & \multirow{2}{*}{\makecell[l]{WFm\\[-0.5ex](3px B)}} \\
        \cline{3-4}
          &  & IoU & F1-score &  &  \\ \hline
        \noalign{\vskip 0.7pt}
        DeepLabV3+ & ResNet-50 & 86.15 & 92.56 & 91.47 & 39.66 \\
        DeepLabV3+$\rightarrow$ours & ResNet-50 & 86.36 & 92.68 & 91.60 & 49.35$_{\text{+9.69}}$ \\
        SegFormer & MIT-B5 & 87.51 & 93.34 & 92.27 & 41.84 \\
        SegFormer$\rightarrow$ours & MIT-B5 & 87.66 & 93.42 & 92.39 & \textbf{49.78}$_{\text{+7.94}}$ \\
        DINOv2 & ViT-B/14 & 84.77 & 91.76 & 90.57 & 33.60 \\
        DINOv2$\rightarrow$ours & ViT-B/14 & 85.85 & 92.39 & 91.27 & 46.73$_{\text{+13.13}}$ \\
        LSK & LSK-s & 87.61 & 93.39 & 92.32 & 43.82 \\
        LSK$\rightarrow$ours & LSK-s & 87.77 & 93.49 & 92.44 & 49.38$_{\text{+5.56}}$ \\
        \noalign{\hrule height 1pt}
      \end{tabular}}
  \endgroup
\end{table}

As presented in Table~\ref{tab:performance_fgfd}, IDGBR significantly enhances the usability of extracted field parcels, notably achieving a maximum WFm improvement of 13.13\% with the DINOv2 baseline (rising from 33.60\% to 46.73\%). For other architectures, IDGBR consistently boosted performance, increasing WFm by 9.69\% on DeepLabV3+, 7.94\% on SegFormer, and 5.56\% on LSK, further demonstrating its robust effectiveness.

\subsubsection{Evaluation on Cross-Phase Building Segmentation}\leavevmode

\begin{table}[htbp]
  \centering
  \caption{Quantitative comparison (\%) of our proposed method against various general segmentation methods on the post-disaster phase of the WHU Building dataset.}
  \label{tab:performance_whu_building}
  \begingroup
    \footnotesize
    \setlength{\tabcolsep}{3pt}
    \renewcommand{\arraystretch}{1}
    \setlength{\dashlinedash}{2pt}
    \setlength{\dashlinegap}{1.5pt}
    \resizebox{\columnwidth}{!}{
      \begin{tabular}{l|l|cc|c @{\hspace{15pt}} l}
        \noalign{\hrule height 1pt}
        \multirow{2}{*}{Method}
          & \multirow{2}{*}{\begin{tabular}[c]{@{}c@{}}DiscModel\\[-0.3ex]Backbone\end{tabular}}
          & \multicolumn{2}{c|}{Building}
          & \multirow{2}{*}{Acc}
          & \multirow{2}{*}{\makecell[l]{WFm\\[-0.5ex](3px B)}} \\
        \cline{3-4}
          &  & IoU & F1-score &  &  \\ \hline
        \noalign{\vskip 0.7pt}
        DeepLabV3+ & ResNet-50 & 93.00 & 96.37 & 98.91 & 68.41 \\
        DeepLabV3+$\rightarrow$ours & ResNet-50 & 94.37 & 97.11 & 99.13 & 74.96$_{\text{+6.55}}$ \\
        SegFormer & MIT-B5 & 93.58 & 96.68 & 99.00 & 71.16 \\
        SegFormer$\rightarrow$ours & MIT-B5 & 94.72 & 97.29 & 99.18 & 75.91$_{\text{+4.75}}$ \\
        DINOv2 & ViT-B/14 & 91.32 & 95.46 & 98.64 & 62.64 \\
        DINOv2$\rightarrow$ours & ViT-B/14 & 93.45 & 96.61 & 98.98 & 71.59$_{\text{+8.95}}$ \\
        LSK & LSK-s & 95.02 & 97.45 & 99.23 & 78.14 \\
        LSK$\rightarrow$ours & LSK-s & 95.23 & 97.56 & 99.26 & \textbf{78.73}$_{\text{+0.59}}$ \\
        \noalign{\hrule height 1pt}
      \end{tabular}}
  \endgroup
\end{table}

\begin{table}[htbp]
  \centering
  \caption{Quantitative comparison (\%) of our proposed method against various general segmentation methods on the post-reconstruction phase of the WHU Building dataset.}
  \label{tab:performance_whu_building_post}
  \begingroup
    \footnotesize
    \setlength{\tabcolsep}{3pt}
    \renewcommand{\arraystretch}{1}
    \setlength{\dashlinedash}{2pt}
    \setlength{\dashlinegap}{1.5pt}
    \resizebox{\columnwidth}{!}{
      \begin{tabular}{l|l|cc|c @{\hspace{15pt}} l}
        \noalign{\hrule height 1pt}
        \multirow{2}{*}{Method}
          & \multirow{2}{*}{\begin{tabular}[c]{@{}c@{}}DiscModel\\[-0.3ex]Backbone\end{tabular}}
          & \multicolumn{2}{c|}{Building}
          & \multirow{2}{*}{Acc}
          & \multirow{2}{*}{\makecell[l]{WFm\\[-0.5ex](3px B)}} \\
        \cline{3-4}
          &  & IoU & F1-score &  &  \\ \hline
        \noalign{\vskip 0.7pt}
        DeepLabV3+ & ResNet-50 & 91.70 & 95.67 & 98.40 & 66.59 \\
        DeepLabV3+$\rightarrow$ours & ResNet-50 & 92.73 & 96.23 & 98.60 & 71.87$_{\text{+5.28}}$ \\
        SegFormer & MIT-B5 & 92.49 & 96.10 & 98.57 & 71.64 \\
        SegFormer$\rightarrow$ours & MIT-B5 & 93.06 & 96.41 & 98.67 & \textbf{73.95}$_{\text{+2.31}}$ \\
        DINOv2 & ViT-B/14 & 90.78 & 95.17 & 98.22 & 61.78 \\
        DINOv2$\rightarrow$ours & ViT-B/14 & 92.84 & 96.29 & 98.62 & 70.63$_{\text{+8.85}}$ \\
        LSK & LSK-s & 91.81 & 95.73 & 98.43 & 70.81 \\
        LSK$\rightarrow$ours & LSK-s & 91.93 & 95.79 & 98.45 & 71.18$_{\text{+0.37}}$ \\
        \noalign{\hrule height 1pt}
      \end{tabular}}
  \endgroup
\end{table}

In the initial post-disaster phase (Table~\ref{tab:performance_whu_building}), IDGBR yielded significant improvements across all baseline models in every evaluation metric. For DeepLabV3+, SegFormer, and DINOv2, the WFm increased by 6.55\%, 4.75\%, and 8.95\%, respectively. IDGBR maintained stable performance in cross-phase testing (Table~\ref{tab:performance_whu_building_post}). Taking DeepLabV3+ as an example, its WFm improved from 68.41\% to 74.96\% (+6.55\%) and IoU rose from 93.00\% to 94.37\% in the post-disaster phase; while maintaining 71.87\% WFm (+5.28\%) and achieving 92.73\% IoU (from 91.70\% baseline) in the post-reconstruction phase. In terms of overall performance, IDGBR yielded average improvements of 5.21\% in WFm and 1.21\% in IoU during the post-disaster phase, while achieving gains of 4.20\% (WFm) and 0.95\% (IoU) in the post-reconstruction phase, which demonstrates IDGBR's robust cross-temporal adaptability.

\subsection{Evaluation on Multi-Class Segmentation Tasks}

In multi-class segmentation tasks, we employ Potsdam and Vaihingen as benchmark datasets to conduct a comparative analysis of model performance in complex boundary scenarios.

As presented in Table~\ref{tab:performance_Potsdam}, on the Potsdam dataset, the IDGBR framework demonstrates superior boundary refinement by synergistically integrating the semantic discriminability of baseline models with the edge delineation capabilities of diffusion models. Quantitatively, IDGBR yields significant WFm improvements across all tested architectures, with gains ranging from 2.88\% to 8.12\%. Notably, the DINOv2 baseline exhibits the most remarkable enhancement, where our method elevates its WFm score from 35.12\% to 43.24\%.

Similarly, as shown in Table~\ref{tab:performance_Vaihingen}, results on Vaihingen further confirm the effectiveness of the proposed method. Compared with baseline models, our framework effectively improves the WFm metric across different model architectures. For instance, IDGBR yielded WFm improvements of 4.76\%, 1.28\%, 1.28\%, and 2.40\% for DeepLabV3+, SegFormer, DINOv2, and LSK, respectively. These results indicate that IDGBR significantly enhances boundary representation without compromising overall segmentation quality.

\begin{table}[htbp]
\centering
\caption{Quantitative comparison (\%) of our proposed method against various general segmentation methods on the Potsdam dataset.}
\label{tab:performance_Potsdam}
\renewcommand{\arraystretch}{1}
\setlength{\tabcolsep}{6pt}
\setlength{\dashlinedash}{2pt}
\setlength{\dashlinegap}{1.5pt}
\small
  \begin{tabular}{l|l|cc @{\hspace{12pt}} l}
  \noalign{\hrule height 1pt}
  \multirow{2}{*}{Method} & \multirow{2}{*}{\begin{tabular}[c]{@{}c@{}}DiscModel\\[-0.5ex]Backbone\end{tabular}} & \multirow{2}{*}{mean F1} & \multirow{2}{*}{mIoU} & \multirow{2}{*}{\makecell[l]{WFm\\[-0.3ex](3px B)}} \\
   &  &  &  &  \\
  \hline
  \noalign{\vskip 0.7pt}
  DeepLabV3+ & ResNet-50 & 90.87 & 83.53 & 37.40 \\
  DeepLabV3+$\rightarrow$ours & ResNet-50 & 90.76 & 83.37 & 42.14$_{\text{+4.74}}$ \\
  SegFormer & MIT-B5 & 90.67 & 83.15 & 39.71 \\
  SegFormer$\rightarrow$ours & MIT-B5 & 90.78 & 83.36 & 42.59$_{\text{+2.88}}$ \\
  DINOv2 & ViT-B/14 & 90.41 & 82.74 & 35.12 \\
  DINOv2$\rightarrow$ours & ViT-B/14 & 90.61 & 83.12 & \textbf{43.24}$_{\text{+8.12}}$ \\
  LSK & LSK-s & 90.08 & 82.22 & 38.28 \\
  LSK$\rightarrow$ours & LSK-s & 90.33 & 82.65 & 42.76$_{\text{+4.48}}$ \\
  \noalign{\hrule height 1pt}
  \end{tabular}
\end{table}

\begin{table}[htbp]
\centering
\caption{Quantitative comparison (\%) of our proposed method against various general segmentation methods on the Vaihingen dataset.}
\label{tab:performance_Vaihingen}
\renewcommand{\arraystretch}{1}
\setlength{\tabcolsep}{6pt} 
\setlength{\dashlinedash}{2pt}
\setlength{\dashlinegap}{1.5pt}
\small
  \begin{tabular}{l|l|cc @{\hspace{12pt}} l}
  \noalign{\hrule height 1pt}
  \multirow{2}{*}{Method} & \multirow{2}{*}{\begin{tabular}[c]{@{}c@{}}DiscModel\\[-0.5ex]Backbone\end{tabular}} & \multirow{2}{*}{mean F1} & \multirow{2}{*}{mIoU} & \multirow{2}{*}{\makecell[l]{WFm\\[-0.3ex](3px B)}} \\
   &  &  &  &  \\
  \hline
  \noalign{\vskip 0.7pt}
  DeepLabV3+ & ResNet-50 & 85.93 & 75.76 & 40.38 \\
  DeepLabV3+$\rightarrow$ours & ResNet-50 & 86.79 & 77.03 & \textbf{45.14}$_{\text{+4.76}}$ \\
  SegFormer & MIT-B5 & 87.38 & 77.95 & 43.76 \\
  SegFormer$\rightarrow$ours & MIT-B5 & 87.52 & 78.17 & 45.04$_{\text{+1.28}}$ \\
  DINOv2 & ViT-B/14 & 85.45 & 75.09 & 37.88 \\
  DINOv2$\rightarrow$ours & ViT-B/14 & 85.77 & 75.54 & 39.16$_{\text{+1.28}}$ \\
  LSK & LSK-s & 87.04 & 77.39 & 42.62 \\
  LSK$\rightarrow$ours & LSK-s & 87.30 & 77.82 & 45.02$_{\text{+2.40}}$ \\
  \noalign{\hrule height 1pt}
  \end{tabular}
\end{table}

\section{Discussion}

\subsection{Comparative Analysis with Specialized Boundary Optimization Methods}

To rigorously benchmark the boundary refinement capability of IDGBR, we conducted a comprehensive comparison against representative boundary optimization methods. Specifically, comparisons are made against two categories: (1). Discriminative boundary optimization methods, including general vision models (PointRend~\cite{kirillovPointRendImageSegmentation2020}, Mask2Former~\cite{chengMaskedAttentionMaskTransformer2022}) and the remote-sensing-specific MSEONet~\cite{huangMultiscaleSemanticSegmentation2025}; and (2). Diffusion-inspired discriminative segmentation methods, represented by DDP~\cite{jiDDPDiffusionModel2023}. Under this framework, IDGBR is deployed to refine the rough labels generated by SegFormer. Extensive experiments across the aforementioned three binary and two multi-class datasets substantiate that IDGBR yields significant performance margins, even against these methods explicitly engineered for boundary optimization.

\begin{table}[htbp]
\centering
\caption{Quantitative results (\%) of methods specialized for boundary optimization on the CHN6-CUG road dataset.}
\label{tab:optimization_road}
\begingroup
\footnotesize
\setlength{\tabcolsep}{3pt}
\renewcommand{\arraystretch}{1}
\setlength{\dashlinedash}{2pt}
\setlength{\dashlinegap}{1.5pt}
\resizebox{0.7\columnwidth}{!}{
\begin{tabular}{l|cc|c @{\hspace{6pt}} c}
\noalign{\hrule height 1pt}
\multirow{2}{*}{Method}
& \multicolumn{2}{c|}{Road}
& \multirow{2}{*}{Acc}
& \multirow{2}{*}{\makecell[c]{WFm\\[-0.5ex](3px B)}} \\
\cline{2-3}
& IoU & F1-score & & \\ \hline
\noalign{\vskip 0.7pt}
PointRend & 62.79 & 77.11 & 97.45 & 35.36 \\
Mask2Former & 63.88 & 77.96 & 97.52 & 37.61 \\
MSEONet & 61.69 & 76.31 & 97.41 & 31.92 \\
DDP & 64.19 & 78.19 & 97.59 & 36.52 \\
\hdashline
\noalign{\vskip 1.5pt}
IDGBR (ours) & 64.54 & 78.45 & 97.50 & \textbf{37.64} \\
\noalign{\hrule height 1pt}
\end{tabular}}
\endgroup
\end{table}

\begin{table}[htbp]
  \centering
  \caption{Quantitative results (\%) of methods specialized for boundary optimization on the FGFD.}
  \label{tab:optimization_fgfd}
  \begingroup
    \footnotesize
    \setlength{\tabcolsep}{3pt}
    \renewcommand{\arraystretch}{1}
    \setlength{\dashlinedash}{2pt}
    \setlength{\dashlinegap}{1.5pt}
    \resizebox{0.7\columnwidth}{!}{
      \begin{tabular}{l|cc|c @{\hspace{6pt}} c}
        \noalign{\hrule height 1pt}
        \multirow{2}{*}{Method}
          & \multicolumn{2}{c|}{Farmland}
          & \multirow{2}{*}{Acc}
          & \multirow{2}{*}{\makecell[c]{WFm\\[-0.5ex](3px B)}} \\
        \cline{2-3}
          & IoU & F1-score &  &  \\ \hline
        \noalign{\vskip 0.7pt}
        PointRend & 81.86 & 92.84 & 91.66 & 38.76 \\
        Mask2Former & 85.94 & 92.44 & 91.27 & 44.27 \\
        MSEONet & 87.06 & 93.08 & 91.93 & 35.13 \\
        DDP & 87.44 & 93.30 & 92.16 & 42.40 \\
        \hdashline
        \noalign{\vskip 1.5pt}
        IDGBR (ours) & 87.66 & 93.42 & 92.39 & \textbf{49.78} \\
        \noalign{\hrule height 1pt}
      \end{tabular}}
  \endgroup
\end{table}

\begin{table}[htbp]
  \centering
  \caption{Quantitative results (\%) of methods specialized for boundary optimization on the post-disaster phase of the WHU Building dataset.}
  \label{tab:optimization_whu_building}
  \begingroup
    \footnotesize
    \setlength{\tabcolsep}{3pt}
    \renewcommand{\arraystretch}{1}
    \setlength{\dashlinedash}{2pt}
    \setlength{\dashlinegap}{1.5pt}
    \resizebox{0.7\columnwidth}{!}{
      \begin{tabular}{l|cc|c @{\hspace{6pt}} c}
        \noalign{\hrule height 1pt}
        \multirow{2}{*}{Method}
          & \multicolumn{2}{c|}{Building}
          & \multirow{2}{*}{Acc}
          & \multirow{2}{*}{\makecell[c]{WFm\\[-0.5ex](3px B)}} \\
        \cline{2-3}
          & IoU & F1-score &  &  \\ \hline
        \noalign{\vskip 0.7pt}
        PointRend & 92.95 & 96.34 & 98.90 & 68.44 \\
        Mask2Former & 93.51 & 96.65 & 98.99 & 71.33 \\
        MSEONet & 90.36 & 94.94 & 98.48 & 58.34 \\
        DDP  & 93.03 & 96.39 & 98.91 & 69.44 \\
        \hdashline
        \noalign{\vskip 1.5pt}
        IDGBR (ours) & 94.72 & 97.29 & 99.18 & \textbf{75.91}\\
        \noalign{\hrule height 1pt}
      \end{tabular}}
  \endgroup
\end{table}

\begin{table}[htbp]
  \centering
  \caption{Quantitative results (\%) of methods specialized for boundary optimization on the post-reconstruction phase of the WHU Building dataset.}
  \label{tab:optimization_whu_building_post}
  \begingroup
    \footnotesize
    \setlength{\tabcolsep}{3pt}
    \renewcommand{\arraystretch}{1}
    \setlength{\dashlinedash}{2pt}
    \setlength{\dashlinegap}{1.5pt}
    \resizebox{0.7\columnwidth}{!}{
      \begin{tabular}{l|cc|c @{\hspace{6pt}} c}
        \noalign{\hrule height 1pt}
        \multirow{2}{*}{Method}
          & \multicolumn{2}{c|}{Building}
          & \multirow{2}{*}{Acc}
          & \multirow{2}{*}{\makecell[c]{WFm\\[-0.5ex](3px B)}} \\
        \cline{2-3}
          & IoU & F1-score &  &  \\ \hline
        \noalign{\vskip 0.7pt}
        PointRend & 91.16 & 95.37 & 98.28 & 66.99 \\
        Mask2Former & 92.16 & 95.92 & 98.49 & 70.98 \\
        MSEONet & 88.34 & 93.81 & 97.72 & 56.67 \\
        DDP  & 92.97 & 96.35 & 98.64 & 72.23 \\
        \hdashline
        \noalign{\vskip 1.5pt}
        IDGBR (ours) & 93.06 & 96.41 & 98.67 & \textbf{73.95} \\
        \noalign{\hrule height 1pt}
      \end{tabular}}
  \endgroup
\end{table}

In binary segmentation scenarios, IDGBR consistently demonstrates superior boundary refinement across datasets with diverse characteristics. As evidenced in Table~\ref{tab:optimization_road}, it effectively outperforms boundary optimization methods on the CHN6-CUG road dataset. This advantage is further amplified on the FGFD dataset (Table~\ref{tab:optimization_fgfd}), where severe spectral ambiguity exists along boundaries. Here, IDGBR achieves the best IoU (87.66\%) and F1-score (93.42\%), while its WFm (49.78\%) substantially leads all competitors, demonstrating the robustness of our method in boundary optimization. Furthermore, IDGBR exhibits strong cross‑temporal generalization on the WHU Building dataset (Tables~\ref{tab:optimization_whu_building} and~\ref{tab:optimization_whu_building_post}). In particular, in the challenging post‑disaster phase, it attains a WFm of 75.91\%, surpassing the best discriminative baseline Mask2Former by over 4.5\%.

\begin{table}[htbp]
\centering
\caption{Quantitative results (\%) of methods specialized for boundary optimization on the Potsdam dataset.}
\label{tab:optimization_Potsdam}
\renewcommand{\arraystretch}{1}
\setlength{\tabcolsep}{6pt}
\setlength{\dashlinedash}{2pt}
\setlength{\dashlinegap}{1.5pt}
\footnotesize
\resizebox{0.75\columnwidth}{!}{
  \begin{tabular}{l|cc @{\hspace{12pt}} c}
  \noalign{\hrule height 1pt}
  \multirow{2}{*}{Method} & \multirow{2}{*}{mean F1} & \multirow{2}{*}{mIoU} & \multirow{2}{*}{\makecell[c]{WFm\\[-0.3ex](3px B)}} \\
   &  &  &  \\
  \hline
  \noalign{\vskip 0.7pt}
  PointRend & 90.56 & 83.00 & 35.87 \\
  Mask2Former & 91.26 & 84.21 & 40.79 \\
  MSEONet & 90.05 & 82.17 & 34.28 \\
  DDP & 90.78 & 83.36 & 41.11 \\
  \hdashline
  \noalign{\vskip 1.5pt}
  IDGBR (ours) & 90.78 & 83.36 & \textbf{42.59} \\
  \noalign{\hrule height 1pt}
  \end{tabular}
}
\end{table}

\begin{table}[htbp]
\centering
\caption{Quantitative results (\%) of methods specialized for boundary optimization on the Vaihingen dataset.}
\label{tab:optimization_Vaihingen}
\renewcommand{\arraystretch}{1}
\setlength{\tabcolsep}{6pt}
\setlength{\dashlinedash}{2pt}
\setlength{\dashlinegap}{1.5pt}
\footnotesize
\resizebox{0.75\columnwidth}{!}{
  \begin{tabular}{l|cc @{\hspace{12pt}} c}
  \noalign{\hrule height 1pt}
  \multirow{2}{*}{Method} & \multirow{2}{*}{mean F1} & \multirow{2}{*}{mIoU} & \multirow{2}{*}{\makecell[c]{WFm\\[-0.3ex](3px B)}} \\
   &  &  &  \\
  \hline
  \noalign{\vskip 0.7pt}
  PointRend & 86.57 & 76.68 & 41.74 \\
  Mask2Former & 86.72 & 76.88 & 43.23 \\
  MSEONet & 86.12 & 76.04 & 39.71 \\ 
  DDP & 86.42 & 76.51 & \textbf{45.05} \\
  \hdashline
  \noalign{\vskip 1.5pt}
  IDGBR (ours) & 87.52 & 78.17 & 45.04 \\
  \noalign{\hrule height 1pt}
  \end{tabular}
}
\end{table}

In multi-class segmentation scenarios, IDGBR maintains a significant advantage in boundary optimization. As presented in Tables~\ref{tab:optimization_Potsdam} and~\ref{tab:optimization_Vaihingen}, on the Vaihingen dataset, our method achieves a WFm of 45.04\%, comparable to the top-performing DDP. Meanwhile, on the Potsdam dataset, it attains a leading WFm of 42.59\%.

\subsection{Enhancing Semantic Awareness in Diffusion Inference: Effects of Conditional Guidance and Cubic Timestep Sampling}

\begin{table}[htbp]
\centering
\caption{Quantitative results (\%) of our framework under different guidance conditions on CHN6-CUG and Potsdam datasets.}
\label{tab:table8}
\footnotesize 
\setlength{\tabcolsep}{1.5pt} 
\begin{tabular}{ccc;{2pt/2pt}ccc;{2pt/2pt}ccc}
\toprule
\multicolumn{3}{c;{2pt/2pt}}{Requirements} & \multicolumn{3}{c;{2pt/2pt}}{CHN6-CUG} & \multicolumn{3}{c}{Potsdam} \\
\cmidrule(r){1-3} \cmidrule(lr){4-6} \cmidrule(l){7-9} 
Image & Rough & Cubic & mean F1 & mIoU & \makecell{WFm\\[-0.5ex](3px B)} & mean F1 & mIoU & \makecell{WFm\\[-0.5ex](3px B)} \\
\midrule
\ding{51} & \ding{55} & \ding{55} & 85.34 & 76.66 & 33.66 & 73.89 & 58.90 & 30.39 \\
\ding{51} & \ding{51} & \ding{55} & \textbf{88.56} & \textbf{80.96} & \textbf{37.64} & 89.29 & 80.88 & 40.34 \\
\ding{51} & \ding{51} & \ding{51} & 87.96 & 80.14 & 37.16 & \textbf{90.78} & \textbf{83.36} & \textbf{42.59} \\
\bottomrule
\end{tabular}
\end{table}

This section mainly discusses the applicability and limitations of diffusion models in semantic segmentation, with analysis conducted at two levels: (1). Exploring the role of conditional guidance in category discrimination and boundary refinement in diffusion models; (2). Exploring the role of cubic timestep sampling mechanism to improve overall performance and semantic consistency. Specifically, the experiments use SegFormer as the discriminative baseline model and are conducted on both the Potsdam and Vaihingen datasets.

In the proposed IDGBR framework, we use remote sensing images and rough labels as conditional guidance, and introduce a cubic timestep sampling strategy to enhance semantic representation capabilities. To systematically assess the contribution of each component, we designed three ablation experiments, which can be divided into three groups: (1). Experiment A: using only remote sensing images as guidance; (2). Experiment B: using remote sensing images and rough labels jointly guided by the proposed Conditional Guidance Module; (3). Experiment C: introducing cubic timestep sampling technique into the timestep sampling based on Experiment B. As shown in Table~\ref{tab:table8}, we compared three different strategies: \enquote{Image} refers to the use of remote sensing image guidance, \enquote{Rough} refers to the use of rough label guidance, and \enquote{Cubic} refers to the additional use of the cubic timestep sampling technique; check marks and crosses are used to indicate whether the corresponding feature is enabled.

By comparing the three sets of experimental results, Experiment A recorded the lowest WFm values (33.66\% on CHN6-CUG; 30.39\% on Potsdam) along with significantly reduced mF1 and mIoU versus other groups. This indicates that without rough label guidance, the model struggles to capture object boundaries. In Experiment B, WFm of the two datasets increased to 37.64\% and 40.34\% respectively, and the mIoU also improved substantially. This demonstrates that rough labels as additional semantic guidance play a crucial role in boundary modeling. In Experiment C, WFm in the Potsdam dataset further increased to 42.59\%, with other metrics also showing an upward trend. This indicates that cubic timestep sampling strengthens feature discrimination at semantic boundaries across multiple classes. However, on the CHN6-CUG binary segmentation dataset, this strategy resulted in a slight decline across all evaluation metrics. Our extended analysis in the Appendix indicates that the cubic sampling strategy is best viewed as a condition-sensitive, optional optimization technique, particularly for binary segmentation.

\subsection{Discussion on Applicability and Limitations}

To quantify the model's repair efficacy on rough predictions, we constructed a partition-based evaluation metric derived from the discrepancy between the rough prediction $M_r$ and the ground truth $M_t$. First, we partition the rough prediction space into the mask of True predicted Region $M_{RTm}$ and the mask of False predicted Region $M_{RFm}$:
\begin{equation}
\begin{aligned}
M_{RTm} &= M_t \cap M_r \\
M_{RFm} &= (M_t \cup M_r) - (M_t \cap M_r)
\end{aligned}
\end{equation}
Based on this partition, we calculate the IoU of the refined prediction $M_{re}$ within these two specific regions:
\begin{equation}
\begin{aligned}
IoU_{RTm} &= \frac{|M_t \cap M_{re} \cap M_{RTm}|}{|(M_t \cap M_{RTm}) \cup (M_{re} \cap M_{RTm})|} \\
IoU_{RFm} &= \frac{|M_t \cap M_{re} \cap M_{RFm}|}{|(M_t \cap M_{RFm}) \cup (M_{re} \cap M_{RFm})|}
\end{aligned}
\end{equation}
where $IoU_{RTm}$ measures the model's capability to preserve true predicted regions; an excessively low value indicates significant degradation of correct regions. $IoU_{RFm}$ evaluates the model's ability to rectify false predicted regions, where a positive value indicates effective repair. Integrating both preservation and rectification capabilities, we define the final improvement score $IoU_{imp}$ by introducing a penalty term $k$ to balance their weights:
\begin{equation}
IoU_{imp} = IoU_{RFm} - k \times (1 - IoU_{RTm}), \quad k=1
\end{equation}
Based on the value of $IoU_{imp}$, we categorize repair efficacy into three levels:
\begin{itemize}
    \item Error Repair ($IoU_{imp} < 0$): Implies that the damage to correct regions outweighs the repair benefits.
    \item Partial Repair ($0 \le IoU_{imp} < 80\%$): Indicates effective repair of some errors without causing severe damage to correct features.
    \item Complete Repair ($IoU_{imp} \ge 80\%$): Represents efficient correction of erroneous regions while perfectly preserving the features of correct regions.
\end{itemize}

We recalculated the quantitative results for the three study scenarios using the $IoU_{imp}$ metric, applying it specifically to the Regions of Interest (ROI) in the diagrams. The following discussion analyzes and interprets these refined results as presented in the figure.

\begin{figure*}[!t]
\centering
\includegraphics[width=1\textwidth]{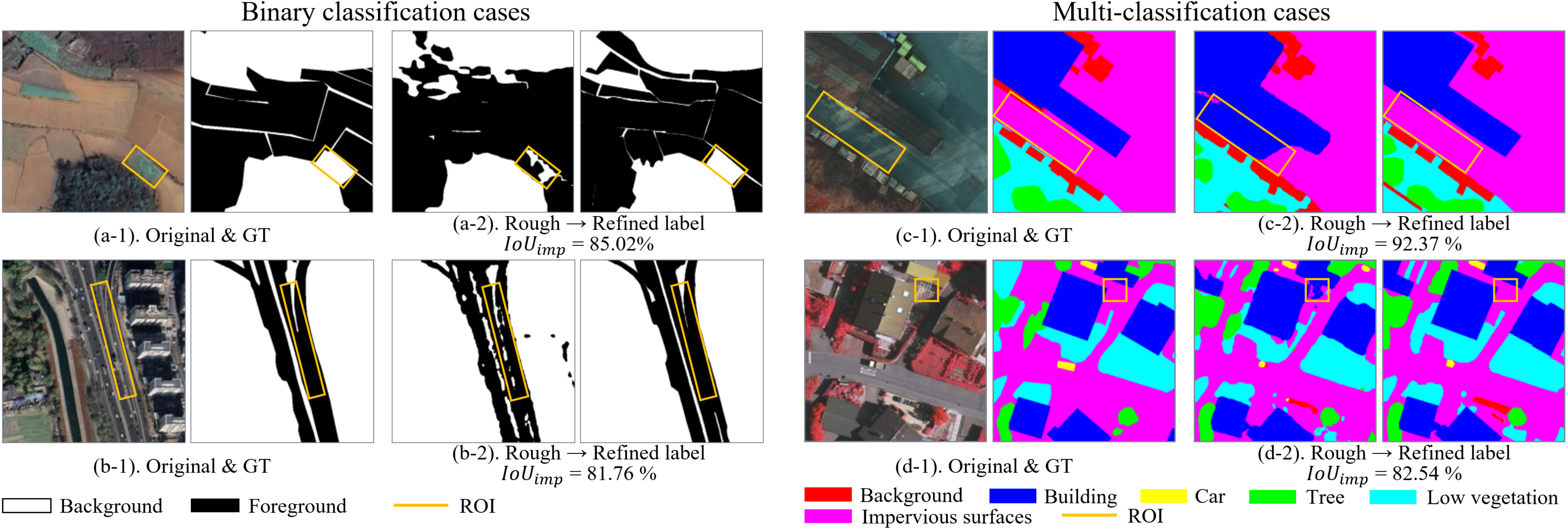}
\caption{Examples of complete semantic patch restoration.}
\label{fig7}
\end{figure*}

When semantic patches exhibit partial segmentation omissions or commissions, yet the correct patch regions can still approximately represent or deduce complete information through semantic adjacency relationships, our method effectively repairs erroneous semantics. Fig.~\ref{fig7} showcases complete semantic patch restoration cases. (i) In (a-2), the refined result rectifies commission errors in the rough label by drawing on learned structure features, achieving an $IoU_{imp}$ of 85.02\% for the ROI. (ii) In (b-2), the spatial position of the road patch is basically correct, but its internal structure appears fragmented. The road continuity patterns learned by the model during training effectively correct the local errors within the patch. (iii) In (c-2), the rough label misclassified impervious surfaces within the ROI as buildings. However, this region exhibits high feature homogeneity and spatial continuity with adjacent correctly labeled impervious surfaces. Leveraging these contextual consistency cues, the refined result effectively rectified the semantic error and regularized the boundaries, achieving a repair efficacy of $IoU_{imp} = 92.37\%$. (iv) In (d-2), the impervious surface patch in the ROI region was misclassified as building categories, but the model corrected these scattered classification errors by utilizing the continuity patterns of impervious surfaces.

\begin{figure*}[!t]
\centering
\includegraphics[width=1\textwidth]{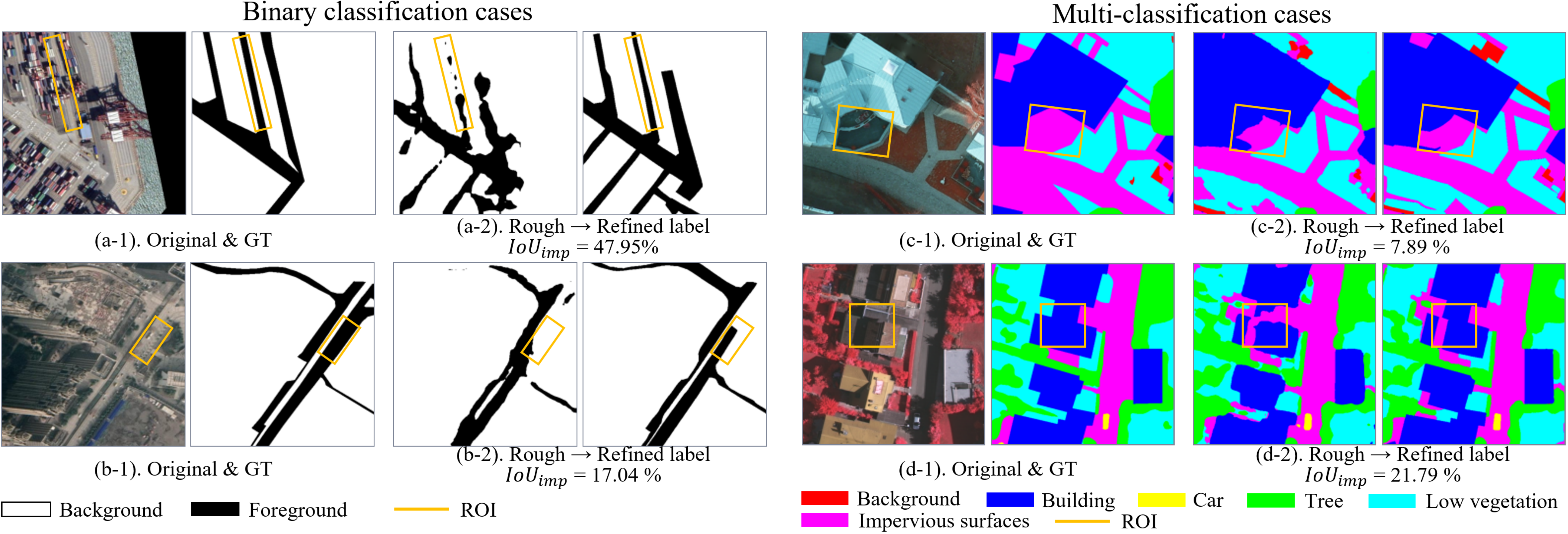}
\caption{Examples of partial semantic patch restoration.}
\label{fig8}
\end{figure*}
When semantic patches exhibit partial segmentation omissions or commissions, yet correct patch regions can only infer localized information through semantic adjacency relations, our method only corrects the relevant local semantics. Fig.~\ref{fig8} showcases partial semantic patch restoration. (i) In (a-2), the rough label exhibits massive segmentation omissions in the ROI road area. Although learned geometric continuity priors successfully connected fragmented patches longitudinally into a complete linear structure, the scarcity of lateral information constrained accurate width inference. Consequently, the optimization achieved topological connectivity repair but failed to fully restore the original width, resulting in a partial repair efficacy of $IoU_{imp} = 47.95\%$. (ii) In (b-2), there are only a few road semantic patches at the bottom of the ROI. The road geometric patterns learned by the model during training can geometrically correct this area but are insufficient to infer the length information of the road. (iii) In (c-2), the impervious surface in the ROI was misclassified as a building. Its high spectral similarity to adjacent buildings prevented semantic correction. Therefore, the model was limited to local geometric refinement. (iv) In (d-2), low-rise buildings on the left side of the ROI region were misclassified as impervious surfaces due to radiometric distortion caused by shadow interference from adjacent high-rise structures on the right. The model could correct the misclassified patches on the right through image and coarse label comparison, but the shadow-affected areas on the left side remained uncorrected.

\begin{figure*}[!t]
\centering
\includegraphics[width=1\textwidth]{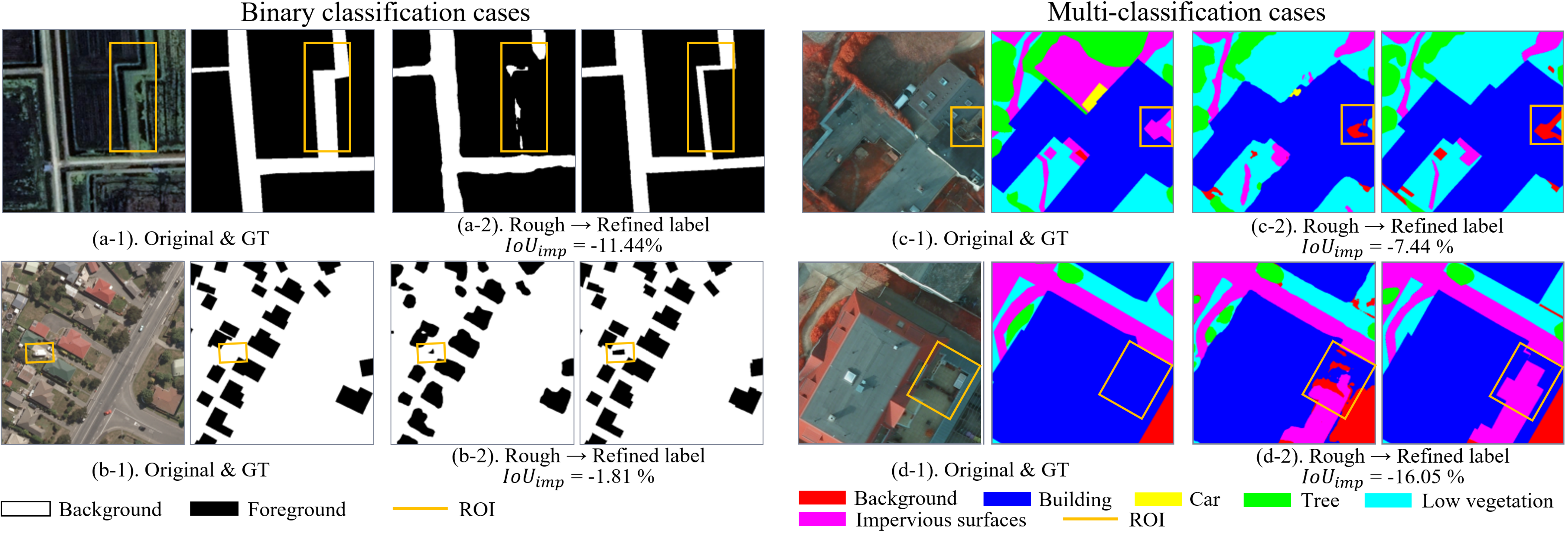}
\caption{Examples of failed semantic patch restoration.}
\label{fig9}
\end{figure*}

When semantic patches exhibit complete omissions or commission errors, and the semantic content cannot be reliably inferred from adjacent patches, our method is unable to rectify these inaccuracies. Fig.~\ref{fig9} showcases instances where incorrect semantic patches cannot be restored. (i) In (a-2), the majority of non-cultivated land within the ROI was misclassified. Although the model repaired topological interruptions based on continuity priors, the lack of localization cues caused the reconstructed trajectory to diverge from the true background and encroach into farmland. This significant spatial deviation resulted in negative optimization, with a final score of $IoU_{imp} = -11.44\%$. (ii) In (b-2), a white vehicle was misclassified due to its spectral similarity to buildings, which made semantic correction very difficult. The model was therefore limited to erroneously refining the contours of the spurious artifact. (iii) In (c-2), impervious surface patches in the ROI were misclassified as background. Despite comparing the image with rough label, the model corrected only the geometric errors while failing to address the semantic misclassification. (iv) In (d-2), building patches in the ROI were misclassified as impervious surfaces. The area shows open-air renovated balconies that closely resemble impervious surfaces in texture and color, causing the model to fail to detect the semantic errors in the rough label, resulting in negative optimization with $IoU_{imp} = -16.05\%$.

\subsection{Speeding Up Representation Learning through Incipient Alignment Strategies}

\begin{figure}[!t]
\centering
\includegraphics[width=3.4in]{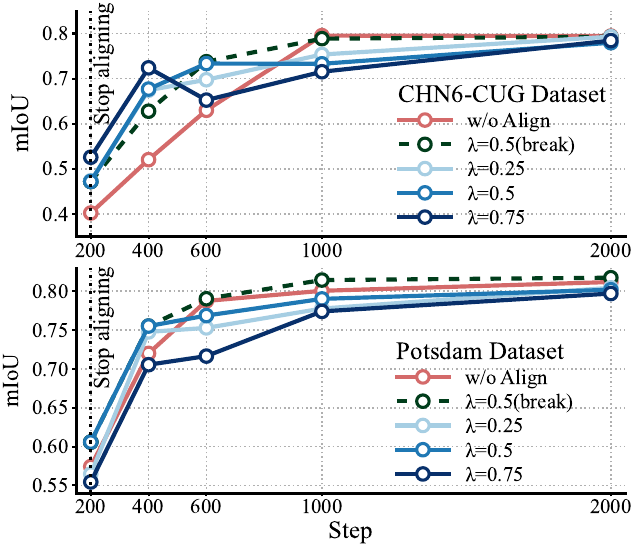}
\caption{The impact of representation alignment mechanism and alignment strength ($\lambda$) on model performance using CHN6-CUG and Potsdam datasets. The w/o Align ($\lambda=0$) group represents training without representation alignment; the break control group stops alignment at step 200.}
\label{fig10}
\end{figure}

To validate the effectiveness of our hyperparameter selection, we analyzed model convergence behaviors under varying alignment strengths ($\lambda \in \{0.25, 0.5, 0.75\}$). As illustrated in Fig.~\ref{fig10}, introducing representation alignment significantly accelerates convergence in the early training stages. However, distinct behavioral patterns emerge as training progresses: excessive regularization (e.g., $\lambda=0.75$) leads to severe performance oscillation and degradation, indicating that overly strong external constraints inhibit the model's intrinsic capacity to capture task-specific semantics. Furthermore, continuous alignment generally yields diminishing returns or even substantial performance declines compared to the non-aligned baseline in later stages (after 400 steps). Consequently, we identified $\lambda=0.5$ combined with an early stopping strategy as the optimal configuration. These results confirm that the alignment mechanism functions best as initial guidance to leverage prior knowledge for accelerated convergence.

\subsection{Effects of Boundary Tolerance Thresholds in WF-measure}

\begin{table}[htbp]
\centering
\caption{Boundary accuracy results of our framework under different boundary thresholds on the Potsdam and Vaihingen datasets\label{tab:table9}}
\setlength{\tabcolsep}{7pt}
\setlength{\aboverulesep}{0pt} 
\setlength{\belowrulesep}{0pt}
\begin{tabular}{cccc}
\toprule
\multirow{2}{*}{\textbf{Boundary}} & \multirow{2}{*}{\textbf{Method}} & \multicolumn{2}{c}{\textbf{WFm}} \\ 
\cmidrule(lr){3-4} 
 & & \textbf{Potsdam} & \textbf{Vaihingen} \\ 
\midrule 
\multirow{2}{*}{1px} & SegFormer & 26.92 & 28.46 \\
 & SegFormer $\rightarrow$ IDGBR & 29.75 (+2.83) & 29.65 (+1.19) \\
\multirow{2}{*}{3px} & SegFormer & 39.71 & 43.76 \\
 & SegFormer $\rightarrow$ IDGBR & 42.59 (+2.88) & 45.04 (+1.28) \\
\multirow{2}{*}{5px} & SegFormer & 47.42 & 53.24 \\
 & SegFormer $\rightarrow$ IDGBR & 49.90 (+2.48) & 54.36 (+1.12) \\
\bottomrule
\end{tabular}
\end{table}

To evaluate the impact of boundary thresholds in the WF-measure metric, boundary thresholds were set at three different values: 1 pixel (denoted as 1px), 3 pixels (denoted as 3px), and 5 pixels (denoted as 5px). The boundary accuracy of the SegFormer model was calculated on both the Potsdam and Vaihingen datasets, with results shown below:

The results (shown in Table~\ref{tab:table9}) demonstrate that as the threshold increases, the WFm metric exhibits an overall upward trend, which aligns with the expectation that higher tolerance relaxes boundary matching requirements. Meanwhile, IDGBR maintains stable accuracy gains across all thresholds, with relatively small variations in improvement magnitude at different threshold levels (approximately 2.5\%–2.9\% for Potsdam and 1.1\%–1.3\% for Vaihingen). This indicates that IDGBR framework is insensitive to changes in boundary tolerance and demonstrates consistent boundary accuracy advantages under various boundary threshold conditions, exhibiting robust boundary optimization capabilities.

\section{Conclusion}
In this work, we address the issue of inaccurate boundary delineation in remote sensing semantic segmentation. First, we observe that discriminative learning excels at capturing low-frequency features but exhibits inherent limitations in learning high-frequency features for semantic segmentation. Subsequently, we validate through qualitative experimental analysis and quantitative theoretical proof that the diffusion denoising process significantly enhances the model's capability to learn high-frequency features. We further observe that these models lack sufficient semantic inference capabilities for low-frequency features when solely guided by the original image. Consequently, we integrate the strengths of discriminative learning and diffusion-based generative learning to propose IDGBR. This framework employs a conditional guidance network to learn joint guidance representation from the coarse segmentation map and the input image, thereby leveraging the respective advantages of discriminative and generative models in semantic and boundary aspects. We conducted extensive experiments on CNN-based architectures, Transformer-based architectures, and self-supervised models, alongside comparative benchmarks against boundary optimization methods, covering both binary and multi-class remote sensing semantic segmentation scenarios. The results demonstrate IDGBR’s robust boundary refinement and superiority over boundary optimization methods. We also discuss the crucial roles of conditional guidance and cubic timestep sampling techniques in integrating the advantages of discriminative and generative approaches. Furthermore, we categorize IDGBR's performance into three levels based on the proposed repair efficacy metric, systematically discussing the method's applicability and limitations. Subsequently, experiments analyze the effectiveness of regularization techniques and boundary threshold effects in WF-measure.

Future work can explore two primary directions. First, conditional diffusion models rely on multi-step iterative sampling during inference. While this strategy significantly improves segmentation quality, it also introduces additional computational overhead and inference latency. Therefore, developing more efficient sampling strategies while preserving segmentation accuracy remains a critical challenge. Second, the optimization performance of our method in some scenarios relies on the completeness of prompt information. Future research could further explore the synergistic effects of multi-granularity spatial prompts and language guidance in the optimization process to improve optimization robustness and segmentation accuracy in complex scenarios.

\bibliographystyle{IEEEtran}
\bibliography{a}

\begin{thebibliography}{10}
\providecommand{\url}[1]{#1}
\csname url@samestyle\endcsname
\providecommand{\newblock}{\relax}
\providecommand{\bibinfo}[2]{#2}
\providecommand{\BIBentrySTDinterwordspacing}{\spaceskip=0pt\relax}
\providecommand{\BIBentryALTinterwordstretchfactor}{4}
\providecommand{\BIBentryALTinterwordspacing}{\spaceskip=\fontdimen2\font plus
\BIBentryALTinterwordstretchfactor\fontdimen3\font minus \fontdimen4\font\relax}
\providecommand{\BIBforeignlanguage}[2]{{%
\expandafter\ifx\csname l@#1\endcsname\relax
\typeout{** WARNING: IEEEtran.bst: No hyphenation pattern has been}%
\typeout{** loaded for the language `#1'. Using the pattern for}%
\typeout{** the default language instead.}%
\else
\language=\csname l@#1\endcsname
\fi
#2}}
\providecommand{\BIBdecl}{\relax}
\BIBdecl

\bibitem{heSwinTransformerEmbedding2022}
X.~He, Y.~Zhou, J.~Zhao, D.~Zhang, R.~Yao, and Y.~Xue, ``{Swin} transformer embedding {UNet} for remote sensing image semantic segmentation,'' \emph{IEEE Trans. Geosci. Remote Sens.}, vol.~60, pp. 1--15, 2022.

\bibitem{hamaguchiEffectiveUseDilated2018}
R.~Hamaguchi, A.~Fujita, K.~Nemoto, T.~Imaizumi, and S.~Hikosaka, ``Effective use of dilated convolutions for segmenting small object instances in remote sensing imagery,'' in \emph{Proc. IEEE/CVF Winter Conf. Appl. Comput. Vis.}, 2018, pp. 1442--1450.

\bibitem{chenGeneralizedAsymmetricDualFront2021}
D.~Chen, J.~A. Spencer, J.~Mirebeau, K.~Chen, M.~Shu, and L.~D. Cohen, ``A generalized asymmetric dual-front model for active contours and image segmentation,'' \emph{IEEE Trans. Image Process.}, vol.~30, pp. 5056--5071, 2021.

\bibitem{marmanisClassificationEdgeImproving2018}
D.~Marmanis, K.~Schindler, J.~D. Wegner, S.~Galliani, M.~Datcu, and U.~Stilla, ``Classification with an edge: Improving semantic image segmentation with boundary detection,'' \emph{ISPRS J. Photogramm. Remote Sens.}, vol. 135, pp. 158--172, 2018.

\bibitem{kirillovPointRendImageSegmentation2020}
A.~Kirillov, Y.~Wu, K.~He, and R.~B. Girshick, ``{PointRend}: Image segmentation as rendering,'' in \emph{Proc. IEEE/CVF Conf. Comput. Vis. Pattern Recognit.}, 2020, pp. 9796--9805.

\bibitem{bertasiusConvolutionalRandomWalk2017}
G.~Bertasius, L.~Torresani, S.~X. Yu, and J.~Shi, ``Convolutional random walk networks for semantic image segmentation,'' in \emph{Proc. IEEE Conf. Comput. Vis. Pattern Recognit.}, 2017, pp. 6137--6145.

\bibitem{dingBoundaryAwareFeaturePropagation2019a}
H.~Ding, X.~Jiang, A.~Q. Liu, N.~Magnenat{-}Thalmann, and G.~Wang, ``Boundary-aware feature propagation for scene segmentation,'' in \emph{Proc. IEEE/CVF Int. Conf. Comput. Vis.}, 2019, pp. 6818--6828.

\bibitem{wangGeometricBoundaryGuided2023}
Y.~Wang, H.~Zhang, Y.~Hu, X.~Hu, L.~Chen, and S.~Hu, ``Geometric boundary guided feature fusion and spatial-semantic context aggregation for semantic segmentation of remote sensing images,'' \emph{IEEE Trans. Image Process.}, vol.~32, pp. 6373--6385, 2023.

\bibitem{wuConditionalBoundaryLoss2023}
D.~Wu, Z.~Guo, A.~Li, C.~Yu, C.~Gao, and N.~Sang, ``Conditional boundary loss for semantic segmentation,'' \emph{IEEE Trans. Image Process.}, vol.~32, pp. 3717--3731, 2023.

\bibitem{wangHighFrequencyComponentHelps2020}
H.~Wang, X.~Wu, Z.~Huang, and E.~P. Xing, ``High-frequency component helps explain the generalization of convolutional neural networks,'' in \emph{Proc. IEEE/CVF Conf. Comput. Vis. Pattern Recognit.}, 2020, pp. 8681--8691.

\bibitem{linInvestigatingExplainingFrequency2022}
Z.~Lin, Y.~Gao, and J.~Sang, ``Investigating and explaining the frequency bias in image classification,'' in \emph{Proc. Int. Jt. Conf. Artif. Intell.}, 2022, pp. 717--723.

\bibitem{ardizzoneTrainingNormalizingFlows2020}
L.~Ardizzone, R.~Mackowiak, C.~Rother, and U.~K{\"{o}}the, ``Training normalizing flows with the information bottleneck for competitive generative classification,'' in \emph{Proc. Adv. Neural Inf. Process. Syst.}, 2020, pp. 7828--7840.

\bibitem{ayyoubzadehHighFrequencyDetail2021}
S.~M. Ayyoubzadeh and X.~Wu, ``High frequency detail accentuation in {CNN} image restoration,'' \emph{{IEEE} Trans. Image Process.}, vol.~30, pp. 8836--8846, 2021.

\bibitem{schwarzFrequencyBiasGenerative2021}
K.~Schwarz, Y.~Liao, and A.~Geiger, ``On the frequency bias of generative models,'' in \emph{Proc. Adv. Neural Inf. Process. Syst.}, 2021, pp. 18\,126--18\,136.

\bibitem{rombachHighResolutionImageSynthesis2022a}
R.~Rombach, A.~Blattmann, D.~Lorenz, P.~Esser, and B.~Ommer, ``High-resolution image synthesis with latent diffusion models,'' in \emph{Proc. IEEE/CVF Conf. Comput. Vis. Pattern Recognit.}, 2022, pp. 10\,674--10\,685.

\bibitem{songDenoisingDiffusionImplicit2021}
J.~Song, C.~Meng, and S.~Ermon, ``Denoising diffusion implicit models,'' in \emph{Proc. Int. Conf. Learn. Representations}, 2021, pp. 14\,205--14\,224.

\bibitem{hoDenoisingDiffusionProbabilistic2020}
J.~Ho, A.~Jain, and P.~Abbeel, ``Denoising diffusion probabilistic models,'' in \emph{Proc. Adv. Neural Inf. Process. Syst.}, 2020, pp. 6840--6851.

\bibitem{siFreeUFreeLunch2024}
C.~Si, Z.~Huang, Y.~Jiang, and Z.~Liu, ``{FreeU}: Free lunch in diffusion {U-Net},'' in \emph{Proc. IEEE/CVF Conf. Comput. Vis. Pattern Recognit.}, 2024, pp. 4733--4743.

\bibitem{yuRepresentationAlignmentGeneration2025}
S.~Yu \emph{et~al.}, ``Representation alignment for generation: Training diffusion transformers is easier than you think,'' in \emph{Proc. Int. Conf. Learn. Representations}, 2025, pp. 29\,100--29\,142.

\bibitem{margolinHowEvaluateForeground2014}
R.~Margolin, L.~Zelnik{-}Manor, and A.~Tal, ``How to evaluate foreground maps,'' in \emph{Proc. IEEE Conf. Comput. Vis. Pattern Recognit.}, 2014, pp. 248--255.

\bibitem{DBLP:journals/pami/SuZWZLPL23}
Z.~Su \emph{et~al.}, ``Lightweight pixel difference networks for efficient visual representation learning,'' \emph{IEEE Trans. Pattern Anal. Mach. Intell.}, vol.~45, no.~12, pp. 14\,956--14\,974, 2023.

\bibitem{liSemanticSegmentationRemote2024}
X.~Li, F.~Xu, F.~Liu, Y.~Tong, X.~Lyu, and J.~Zhou, ``Semantic segmentation of remote sensing images by interactive representation refinement and geometric prior-guided inference,'' \emph{IEEE Trans. Geosci. Remote Sens.}, vol.~62, pp. 1--18, 2024.

\bibitem{wuFSVLMVisionLanguageModel2025}
H.~Wu, Z.~Du, D.~Zhong, Y.~Wang, and C.~Tao, ``{FSVLM}: A vision-language model for remote sensing farmland segmentation,'' \emph{IEEE Trans. Geosci. Remote Sens.}, vol.~63, pp. 1--13, 2025.

\bibitem{huangMultiscaleSemanticSegmentation2025}
W.~Huang, F.~Deng, H.~Liu, M.~Ding, and Q.~Yao, ``Multiscale semantic segmentation of remote sensing images based on edge optimization,'' \emph{IEEE Trans. Geosci. Remote Sens.}, vol.~63, pp. 1--13, 2025.

\bibitem{liLSKNetFoundationLightweight2025}
Y.~Li \emph{et~al.}, ``{LSKNet}: A foundation lightweight backbone for remote sensing,'' \emph{Int. J. Comput. Vis.}, vol. 133, no.~3, pp. 1410--1431, 2025.

\bibitem{vaswaniAttentionAllYou2017}
A.~Vaswani \emph{et~al.}, ``Attention is all you need,'' in \emph{Proc. Adv. Neural Inf. Process. Syst.}, 2017, pp. 5998--6008.

\bibitem{zhengRethinkingSemanticSegmentation2021}
S.~Zheng \emph{et~al.}, ``Rethinking semantic segmentation from a sequence-to-sequence perspective with transformers,'' in \emph{Proc. IEEE/CVF Conf. Comput. Vis. Pattern Recognit.}, 2021, pp. 6881--6890.

\bibitem{strudelSegmenterTransformerSemantic2021a}
R.~Strudel, R.~Garcia, I.~Laptev, and C.~Schmid, ``{Segmenter}: Transformer for semantic segmentation,'' in \emph{Proc. IEEE/CVF Int. Conf. Comput. Vis.}, 2021, pp. 7242--7252.

\bibitem{wangUNetFormerUNetlikeTransformer2022}
L.~Wang \emph{et~al.}, ``{{UNetFormer}}: {{A UNet-like}} transformer for efficient semantic segmentation of remote sensing urban scene imagery,'' \emph{ISPRS J. Photogramm. Remote Sens.}, vol. 190, pp. 196--214, 2022.

\bibitem{zhangFsaNetFrequencySelfAttention2023}
F.~Zhang, A.~Panahi, and G.~Gao, ``{FsaNet}: Frequency self-attention for semantic segmentation,'' \emph{IEEE Trans. Image Process.}, vol.~32, pp. 4757--4772, 2023.

\bibitem{borseInverseFormLossFunction2021a}
S.~Borse, Y.~Wang, Y.~Zhang, and F.~Porikli, ``{InverseForm}: A loss function for structured boundary-aware segmentation,'' in \emph{Proc. IEEE/CVF Conf. Comput. Vis. Pattern Recognit.}, 2021, pp. 5901--5911.

\bibitem{wangActiveBoundaryLoss2022}
C.~Wang \emph{et~al.}, ``Active boundary loss for semantic segmentation,'' in \emph{Proc. AAAI Conf. Artif. Intell.}, 2022, pp. 2397--2405.

\bibitem{takikawaGatedSCNNGatedShape2019}
T.~Takikawa, D.~Acuna, V.~Jampani, and S.~Fidler, ``{Gated-SCNN}: Gated shape {CNNs} for semantic segmentation,'' in \emph{Proc. IEEE/CVF Int. Conf. Comput. Vis.}, 2019, pp. 5228--5237.

\bibitem{yinDisentangledNonlocalNeural2020}
M.~Yin \emph{et~al.}, ``Disentangled non-local neural networks,'' in \emph{Proc. Eur. Conf. Comput. Vis.}, 2020, pp. 191--207.

\bibitem{dingLANetLocalAttention2021}
L.~Ding, H.~Tang, and L.~Bruzzone, ``{LANet}: Local attention embedding to improve the semantic segmentation of remote sensing images,'' \emph{IEEE Trans. Geosci. Remote Sens.}, vol.~59, no.~1, pp. 426--435, 2021.

\bibitem{zhengHighOrderSemanticDecoupling2023}
C.~Zheng, J.~Nie, Z.~Wang, N.~Song, J.~Wang, and Z.~Wei, ``High-order semantic decoupling network for remote sensing image semantic segmentation,'' \emph{IEEE Trans. Geosci. Remote Sens.}, vol.~61, pp. 1--15, 2023.

\bibitem{chengMaskedAttentionMaskTransformer2022}
B.~Cheng, I.~Misra, A.~G. Schwing, A.~Kirillov, and R.~Girdhar, ``Masked-attention mask transformer for universal image segmentation,'' in \emph{Proc. IEEE/CVF Conf. Comput. Vis. Pattern Recognit.}, 2022, pp. 1290--1299.

\bibitem{raiMask2AnomalyMaskTransformer2024}
S.~N. Rai, F.~Cermelli, B.~Caputo, and C.~Masone, ``{Mask2Anomaly}: Mask transformer for universal open-set segmentation,'' \emph{IEEE Trans. Pattern Anal. Mach. Intell.}, vol.~46, no.~12, pp. 9286--9302, 2024.

\bibitem{liFrequencyDecouplingNetwork2025}
X.~Li, F.~Xu, A.~Yu, X.~Lyu, H.~Gao, and J.~Zhou, ``A frequency decoupling network for semantic segmentation of remote sensing images,'' \emph{IEEE Trans. Geosci. Remote Sens.}, vol.~63, pp. 1--21, 2025.

\bibitem{geirhosImageNetTrainedCNNs2019}
R.~Geirhos, P.~Rubisch, C.~Michaelis, M.~Bethge, F.~A. Wichmann, and W.~Brendel, ``{ImageNet}-trained {CNNs} are biased towards texture; increasing shape bias improves accuracy and robustness,'' in \emph{Proc. Int. Conf. Learn. Representations}, 2019, pp. 9234--9255.

\bibitem{baranchukLabelEfficientSemanticSegmentation2022}
D.~Baranchuk, A.~Voynov, I.~Rubachev, V.~Khrulkov, and A.~Babenko, ``Label-efficient semantic segmentation with diffusion models,'' in \emph{Proc. Int. Conf. Learn. Representations}, 2022, pp. 15\,633--15\,647.

\bibitem{zhaoUnleashingTexttoImageDiffusion2023}
W.~Zhao, Y.~Rao, Z.~Liu, B.~Liu, J.~Zhou, and J.~Lu, ``Unleashing text-to-image diffusion models for visual perception,'' in \emph{Proc. IEEE/CVF Int. Conf. Comput. Vis.}, 2023, pp. 5706--5716.

\bibitem{jiDDPDiffusionModel2023}
Y.~Ji \emph{et~al.}, ``{DDP}: Diffusion model for dense visual prediction,'' in \emph{Proc. IEEE/CVF Int. Conf. Comput. Vis.}, 2023, pp. 21\,684--21\,695.

\bibitem{khannaDiffusionSat2024}
S.~Khanna \emph{et~al.}, ``{DiffusionSat}: A generative foundation model for satellite imagery,'' in \emph{Proc. Int. Conf. Learn. Representations}, 2024, pp. 19\,539--19\,557.

\bibitem{ZhiMetaEarthGenerativeFoundation2025}
Z.~Yu, C.~Liu, L.~Liu, Z.~Shi, and Z.~Zou, ``{MetaEarth}: A generative foundation model for global-scale remote sensing image generation,'' \emph{IEEE Trans. Pattern Anal. Mach. Intell.}, vol.~47, no.~3, pp. 1764--1781, 2025.

\bibitem{liuText2Earth2025}
C.~Liu, K.~Chen, R.~Zhao, Z.~Zou, and Z.~Shi, ``{Text2Earth}: Unlocking text-driven remote sensing image generation with a global-scale dataset and a foundation model,'' \emph{IEEE Geosci. Remote Sens. Mag.}, vol.~13, no.~3, pp. 238--259, 2025.

\bibitem{dongBuildingBridges2024}
R.~Dong \emph{et~al.}, ``Building bridges across spatial and temporal resolutions: Reference-based super-resolution via change priors and conditional diffusion model,'' in \emph{Proc. IEEE/CVF Conf. Comput. Vis. Pattern Recognit.}, 2024, pp. 27\,674--27\,684.

\bibitem{xiaoEDiffSR2024}
Y.~Xiao, Q.~Yuan, K.~Jiang, J.~He, X.~Jin, and L.~Zhang, ``{EDiffSR}: An efficient diffusion probabilistic model for remote sensing image super-resolution,'' \emph{IEEE Trans. Geosci. Remote Sens.}, vol.~62, pp. 1--14, 2024.

\bibitem{leeSpectrumTranslationRefinement2024}
S.~Lee, S.~Jung, and H.~Seo, ``Spectrum translation for refinement of image generation {(STIG)} based on contrastive learning and spectral filter profile,'' in \emph{Proc. AAAI Conf. Artif. Intell.}, 2024, pp. 2929--2937.

\bibitem{yangDiffusionProbabilisticModel2023}
X.~Yang, D.~Zhou, J.~Feng, and X.~Wang, ``Diffusion probabilistic model made slim,'' in \emph{Proc. IEEE/CVF Conf. Comput. Vis. Pattern Recognit.}, 2023, pp. 22\,552--22\,562.

\bibitem{fieldRelationsStatisticsNatural1987}
D.~J. Field, ``Relations between the statistics of natural images and the response properties of cortical cells,'' \emph{J. Opt. Soc. Amer. A}, vol.~4, no.~12, pp. 2379--2394, 1987.

\bibitem{gaoCOMPOSECrossModalPseudoSiamese2020}
J.~Gao, C.~Xiao, L.~M. Glass, and J.~Sun, ``{COMPOSE}: Cross-modal pseudo-siamese network for patient trial matching,'' in \emph{Proc. ACM SIGKDD Int. Conf. Knowl. Discov. Data Min.}, 2020, pp. 803--812.

\bibitem{oquabDINOv2LearningRobust2024}
M.~Oquab \emph{et~al.}, ``{DINOv2}: Learning robust visual features without supervision,'' \emph{Trans. Mach. Learn. Res.}, vol. 2024, 2024.

\bibitem{mouT2IAdapterLearningAdapters2024}
C.~Mou \emph{et~al.}, ``{T2I-Adapter}: Learning adapters to dig out more controllable ability for text-to-image diffusion models,'' in \emph{Proc. AAAI Conf. Artif. Intell.}, 2024, pp. 4296--4304.

\bibitem{zhuGlobalContextawareBatchindependent2021}
Q.~Zhu \emph{et~al.}, ``A global context-aware and batch-independent network for road extraction from {VHR} satellite imagery,'' \emph{ISPRS J. Photogramm. Remote Sens.}, vol. 175, pp. 353--365, 2021.

\bibitem{liComprehensiveDeepLearningFramework2025}
J.~Li, Y.~Wei, T.~Wei, and W.~He, ``A comprehensive deep-learning framework for fine-grained farmland mapping from high-resolution images,'' \emph{IEEE Trans. Geosci. Remote Sens.}, vol.~63, pp. 1--15, 2025.

\bibitem{jiFullyConvolutionalNetworks2019}
S.~Ji, S.~Wei, and M.~Lu, ``Fully convolutional networks for multisource building extraction from an open aerial and satellite imagery data set,'' \emph{IEEE Trans. Geosci. Remote Sens.}, vol.~57, no.~1, pp. 574--586, 2019.

\bibitem{ISPRS2DSemantic2022}
{ISPRS}, ``{2D} semantic labeling contest - {Potsdam},'' [Online]. Available: \url{https://www.isprs.org/resources/datasets/benchmarks/UrbanSemLab/2d-sem-label-potsdam.aspx}, 2022.

\bibitem{ISPRS2DSemantic2022a}
------, ``{2D} semantic labeling contest - {Vaihingen},'' [Online]. Available: \url{https://www.isprs.org/resources/datasets/benchmarks/UrbanSemLab/2d-sem-label-vaihingen.aspx}, 2022.

\bibitem{chenEncoderDecoderAtrousSeparable2018}
L.~Chen, Y.~Zhu, G.~Papandreou, F.~Schroff, and H.~Adam, ``Encoder-decoder with atrous separable convolution for semantic image segmentation,'' in \emph{Proc. Eur. Conf. Comput. Vis.}, 2018, pp. 833--851.

\bibitem{xieSegFormerSimpleEfficient2021}
E.~Xie, W.~Wang, Z.~Yu, A.~Anandkumar, J.~M. {\'{A}}lvarez, and P.~Luo, ``{SegFormer}: Simple and efficient design for semantic segmentation with transformers,'' in \emph{Proc. Adv. Neural Inf. Process. Syst.}, 2021, pp. 12\,077--12\,090.

\bibitem{fvcore}
{Meta Platforms}, ``fvcore,'' GitHub repository. [Online]. Available: \url{https://github.com/facebookresearch/fvcore}, 2019.

\end{thebibliography}

\newpage
\appendices

\twocolumn[
  \begin{center}
    \textbf{\LARGE Supplementary Materials}\par
    \vspace{0.78cm}
  \end{center}
]

\section*{I. Equation Derivation Details}
\setcounter{subsection}{0}
\label{sec:wiener}
\renewcommand{\theequation}{A\arabic{equation}}
\setcounter{equation}{0}
\subsection{Proof of the Wiener Filter Solution in the Diffusion Probabilistic Model}
Here, we present the general mathematical derivation for treating the diffusion model as a Wiener filter. Assume that $z_0$ is a wide-sense stationary signal and $\epsilon$ is white Gaussian noise with variance $\sigma^2 = 1$. In the Diffusion Probabilistic Model (DPM), the signal and noise are mixed at any arbitrary timestep $t$, and the resulting observation can be expressed as:
\(
z_t = \sqrt{\bar{\alpha}_t}\, z_0 + \sqrt{1-\bar{\alpha}_t}\, \epsilon
\)
To further simplify the model and obtain an explicit closed-form solution, we introduce a linear denoising Wiener filter $h_t$. The objective function for the diffusion model can then be redefined as
\begin{equation}
J_t = \left\| \sqrt{\bar{\alpha}_t}\, z_0 - h_t * z_t \right\|^2
\label{eq.18}
\end{equation}
where $\bar{\alpha}_t$ denotes the scaling factor for the signal component at timestep $t$, which decreases as $t$ increases; the symbol ``$*$'' denotes the convolution operator.

To facilitate analysis and derive the optimal solution for the filter, we convert the above time-domain convolution operation into the frequency domain. According to the convolution theorem, the time-domain convolution $h_t * z_t$ is equivalent to the product in the frequency domain, i.e., $\mathcal{F}[h_t * z_t] = \mathcal{F}[h_t] \cdot \mathcal{F}[z_t]$, where $\mathcal{F}[\cdot]$ denotes the Fourier transform and $\mathcal{H}_t$ represents the corresponding filter in the frequency domain. Thus, the transition from the time-domain convolution to frequency-domain multiplication can be expressed as:
\begin{multline}
\mathcal{H}_t \left( \sqrt{\bar{\alpha}_t} z_0 + \sqrt{1-\bar{\alpha}_t} \mathcal{F}[\epsilon] \right) = \\
\mathcal{F} \left[ h_t * \left( \sqrt{\bar{\alpha}_t} z_0 + \sqrt{1-\bar{\alpha}_t} \epsilon \right) \right]
\label{eq.19}
\end{multline}

Furthermore, the error signal $J_t(f)$ can be written as
\begin{equation}
\mathcal{J}_t(f) = \sqrt{\bar{\alpha}_t}\, \mathcal{Z}_0 - \mathcal{H}_t\left(\sqrt{\bar{\alpha}_t}\, \mathcal{Z}_0 + \sqrt{1-\bar{\alpha}_t}\, \mathcal{F}[\epsilon]\right)
\label{eq.20}
\end{equation}
We obtain the filter by minimizing the mean squared error objective:
\begin{multline}
\min_{h} \mathbb{E} \left[ |\mathcal{J}_t(f)|_2^2 \right]
= \min_{h} \mathbb{E} \left[ \mathcal{J}_t(f)^* \mathcal{J}_t(f) \right] \\
= \min_{h} \mathbb{E} \left[
\left(
    \sqrt{\bar{\alpha}_t} \mathcal{Z}_0
    - \mathcal{H}_t \left( \sqrt{\bar{\alpha}_t} \mathcal{Z}_0
    + \sqrt{1-\bar{\alpha}_t} \mathcal{F}[\epsilon] \right)
\right) \cdot \right. \\
\left.
\left(
    \sqrt{\bar{\alpha}_t} \mathcal{Z}_0
    - \mathcal{H}_t \left( \sqrt{\bar{\alpha}_t} \mathcal{Z}_0
    + \sqrt{1-\bar{\alpha}_t} \mathcal{F}[\epsilon] \right)
\right)
\right]
\label{eq.21}
\end{multline}

where the symbol $^*$ denotes the complex conjugate. To obtain the Wiener filter that minimizes the mean squared error, we take the derivative of the objective function with respect to $\mathcal{H}_t(f)$ and set it to zero. Since $\epsilon$ and $z_0$ are conditionally independent, we have:
\begin{multline}
\frac{\partial \mathbb{E}[|\mathcal{J}_t(f)|_2^2]}{\partial \mathcal{H}_t(f)} =
2 \mathcal{H}_t \left( \bar{\alpha}_t |\mathcal{Z}_0(f)|^2 + (1-\bar{\alpha}_t) |\mathcal{F}[\epsilon]|^2 \right) \\
- 2 \bar{\alpha}_t |\mathcal{Z}_0(f)|^2 = 0
\label{eq.22}
\end{multline}
where $|\mathcal{Z}_0(f)|^2 = \mathbb{E}[\mathcal{Z}_0 \mathcal{Z}_0^*]$, $|\mathcal{F}[\epsilon]|^2 = \mathbb{E}[\mathcal{F}[\epsilon] \mathcal{F}[\epsilon]^*]$. Since the variance of the white noise is $\sigma^2$, $|\mathcal{F}[\epsilon]|^2$ is a constant $\sigma^2$. Simplifying further, we obtain:
\begin{equation}
\mathcal{H}_t^*(f) = \frac{ \bar{\alpha}_t |\mathcal{Z}_0(f)|^2 }{ \bar{\alpha}_t |\mathcal{Z}_0(f)|^2 + (1-\bar{\alpha}_t) \sigma^2 }
\label{eq.23}
\end{equation}
Here, $|\mathcal{Z}_0(f)|^2$ denotes the power spectrum of $z_0$, and $\mathcal{H}_t(f)$ denotes the frequency response of $h_t$. According to the commonly assumed power-law distribution of natural image spectra \cite{fieldRelationsStatisticsNatural1987}, we have $\mathbb{E}[|\mathcal{Z}_0(f)|^2] \approx 1 / f^2$. Thus, the Wiener filter in the frequency domain can be expressed as:
\begin{equation}
\mathcal{H}_t^*(f) \approx \frac{ \bar{\alpha}_t }{ \bar{\alpha}_t + (1-\bar{\alpha}_t) f^2 }
\label{eq.24}
\end{equation}

\section*{II. Network Architecture and Dimensional Specifications}
\setcounter{subsection}{0}
\begin{figure*}[!t]
\centering
\includegraphics[width=1\linewidth]{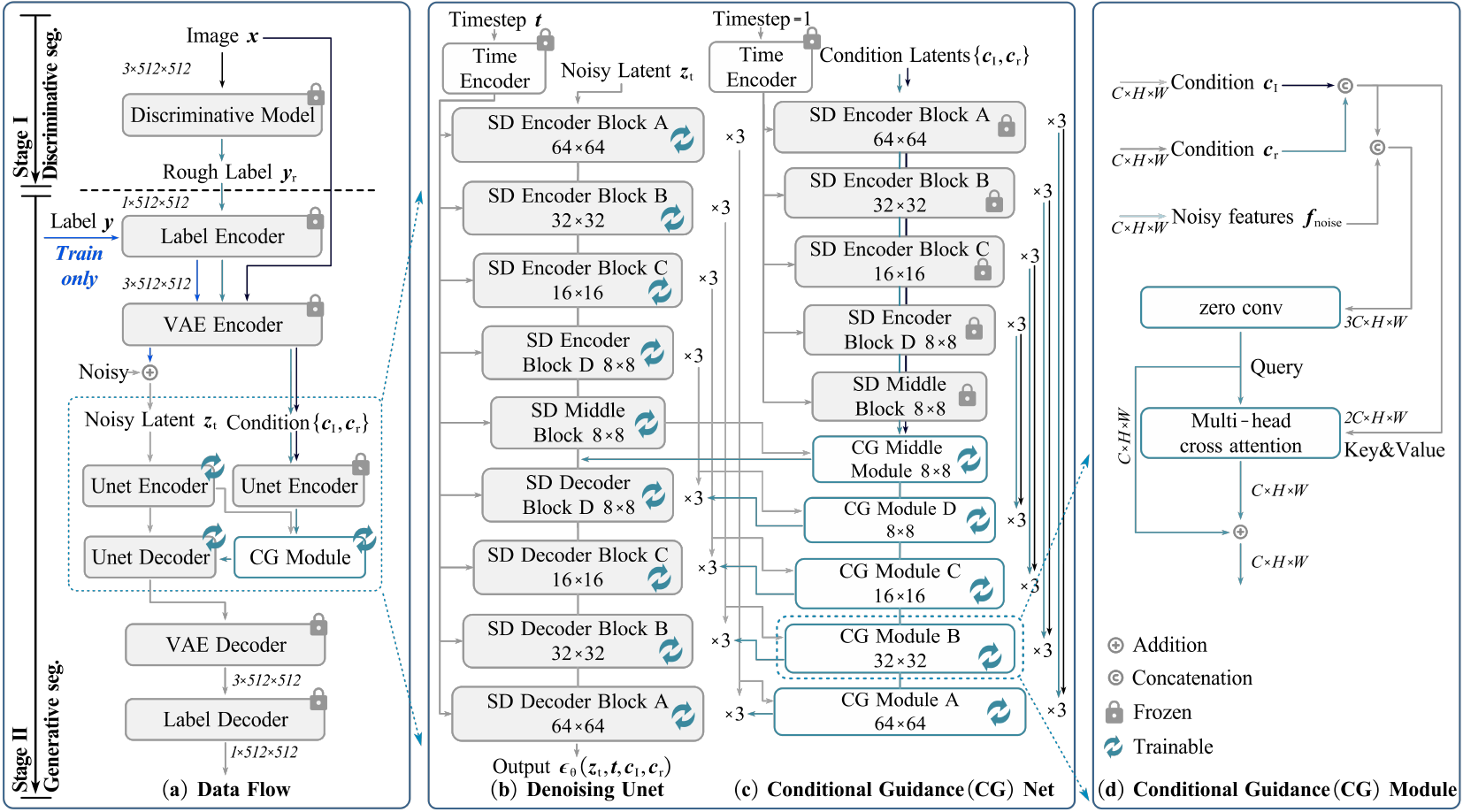}
\caption{Overview of the IDGBR framework. (a) Data Flow: Macroscopically demonstrates the two-stage inference process from the first-stage discriminative coarse segmentation to the second-stage generative fine optimization. (b) Denoising Unet: Illustrates the hierarchical architecture of the backbone network. (c) Conditional Guidance (CG) Net: Highlights the synergistic mechanism between frozen parameters and trainable modules, as well as the feature injection process. (d) Conditional Guidance (CG) Module: Shows the details of the multi-head cross-attention and zero-convolution structures within the module.}
\label{fig:framework_overview}
\end{figure*}

Building upon the schematic diagrams provided in the main text, Fig.~\ref{fig:framework_overview} further details the dimensional specifications and data flow of the proposed IDGBR framework.

Specifically, panel (a) illustrates the collaborative two-stage inference workflow: Stage I utilizes the trained discriminative model to generate a rough label from the original image; in Stage II, the rough label is first fed into the label encoder ($E_e$) to be transformed into continuous features, then concatenated with the original image and compressed by the VAE encoder into latent features to guide the Denoising Unet in generating refined semantic representations, which are ultimately reconstructed into the final high-precision label via the VAE decoder and label decoder ($D_e$). Here, $E_e$ and $D_e$ are introduced to resolve the incompatibility between discrete labels and the continuous latent space. Specifically, $E_e$ maps discrete labels to a 3-channel continuous space via an embedding layer, normalized and scaled to the range $[-1, 1]$ using a Sigmoid function; meanwhile, to enhance training robustness, Gaussian noise with a variance of 0.25 is injected into the features during training before they are fed into the auxiliary decoder $D_e$ (which maps features back to a $K$-channel probability space constrained by cross-entropy loss).

Panels (b) and (c) illustrate in detail the hierarchical encoding processes within the Denoising Unet and the Conditional Guidance (CG) Net. Both networks progressively downsample feature maps from a resolution of $64 \times 64$ to $8 \times 8$. Notably, the encoder of the CG Net serves as a pseudo-Siamese branch of the Denoising Unet encoder, parallelly receiving and encoding the dual conditional inputs $\{c_l, c_r\}$. The extracted multi-scale conditional features are then processed by the CG Module and re-injected into the middle layer and corresponding decoder blocks of the Denoising Unet.

Finally, panel (d) reveals the internal details of the Conditional Guidance (CG) Module. Here, the noisy features $f_{\text{noise}}$ are first concatenated with the conditions $\{c_l, c_r\}$ along the channel dimension to form a hybrid feature map of $3C \times H \times W$, which is processed by a zero-convolution layer to serve as the Query; simultaneously, the dual conditions are concatenated into a $2C \times H \times W$ feature vector to act as the Key and Value for the multi-head cross-attention mechanism. Ultimately, the attention output is fused with the query via a residual connection.

\section*{III. Experimental Details}
\setcounter{subsection}{0}
\subsection{Qualitative Results on the FGFD.}

FGFD is primarily used to validate the boundary optimization capability of the IDGBR framework in complex texture scenes.

\begin{figure*}[!t]
\centering
\includegraphics[width=1\textwidth]{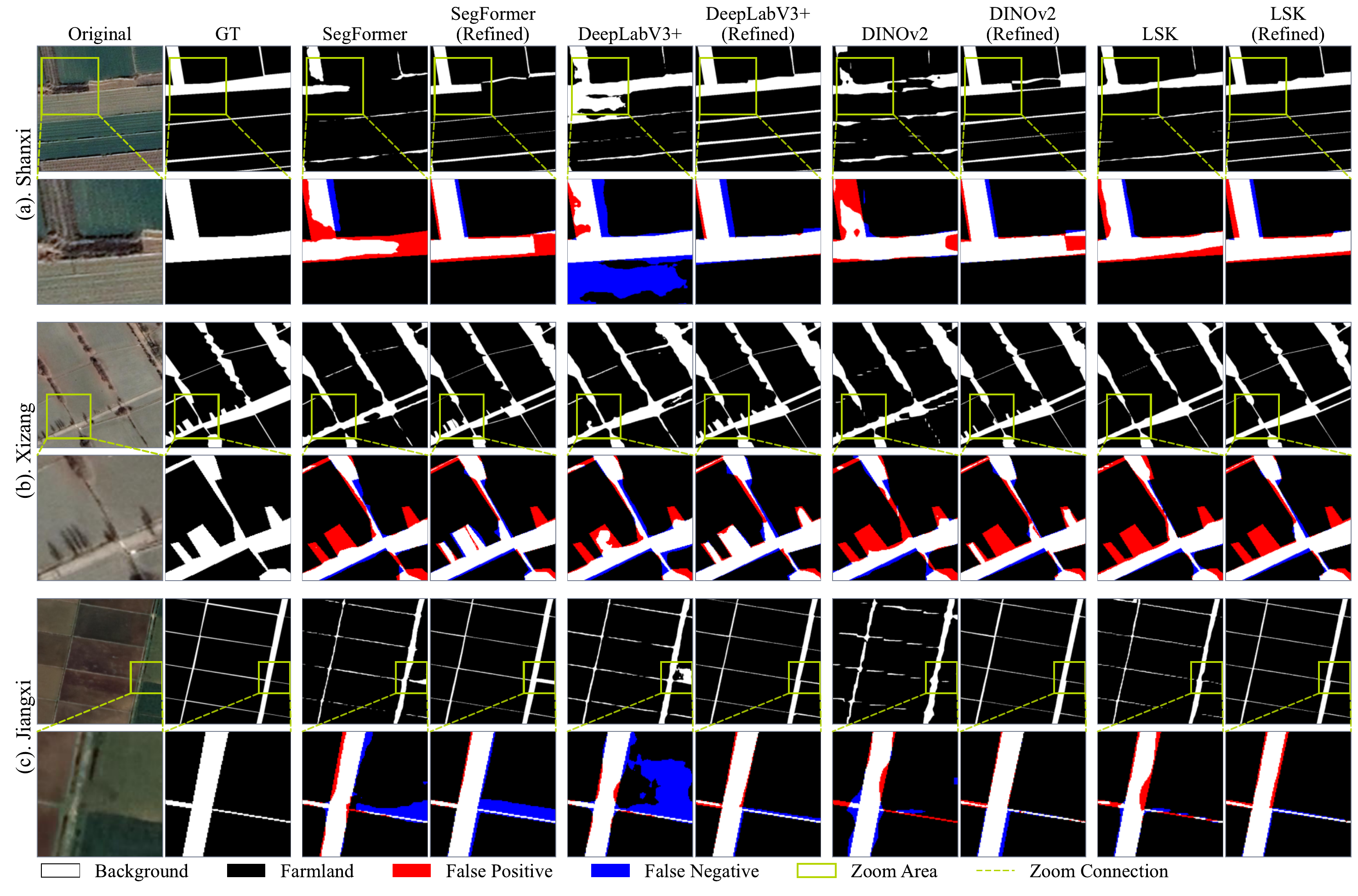}
\caption{Qualitative results of different methods on FGFD.}
\label{fig:supp_fgfd}
\end{figure*}

The complex spatiotemporal attributes of farmland pose significant challenges to traditional discriminative models for segmentation. As shown in Fig.~\ref{fig:supp_fgfd}, the visualization results collectively demonstrate the distinguishability of farmland and the continuity of field roads in three farmland scenarios. Fig.~\ref{fig:supp_fgfd} (a) shows typical northern dryland farmland in Shanxi. For non-cultivated areas that are difficult to distinguish, even when patches appear regular in the original imagery, discriminative models produce patches with incomplete geometric structures and fragmented boundaries. Our method effectively suppresses low-confidence artifact regions, significantly improving patch boundary integrity and standardization. Fig.~\ref{fig:supp_fgfd} (b) depicts highland grassland in Tibet. In the original imagery, shadows cast by trees along farmland boundaries create light-dark divisions within the plots. Discriminative models are prone to boundary misclassification errors in such areas. Our method mitigates segmentation errors of baseline models in shadow regions, enhancing the accuracy and continuity of farmland boundary recognition. Fig.~\ref{fig:supp_fgfd} (c) showcases composite farmland in Jiangxi, where farmland boundaries are often obscured by dense vegetation cover, resulting in blurred boundaries. Due to the minimal spectral differences between vegetation and farmland in the original image, discriminative models struggle to accurately delineate boundary positions, leading to boundary ambiguity and discontinuous segmentation. Our method corrects boundary blurring caused by vegetation interference, improving the geometric consistency and completeness of segmentation results in complex regions.

\subsection{Qualitative Results on the WHU Building Dataset}

To validate the effectiveness of IDGBR in building boundary extraction and cross-temporal generalization, we conducted experiments on the bi-temporal images of the WHU Building Change Detection dataset.

\begin{figure*}[!t]
\centering
\includegraphics[width=1\textwidth]{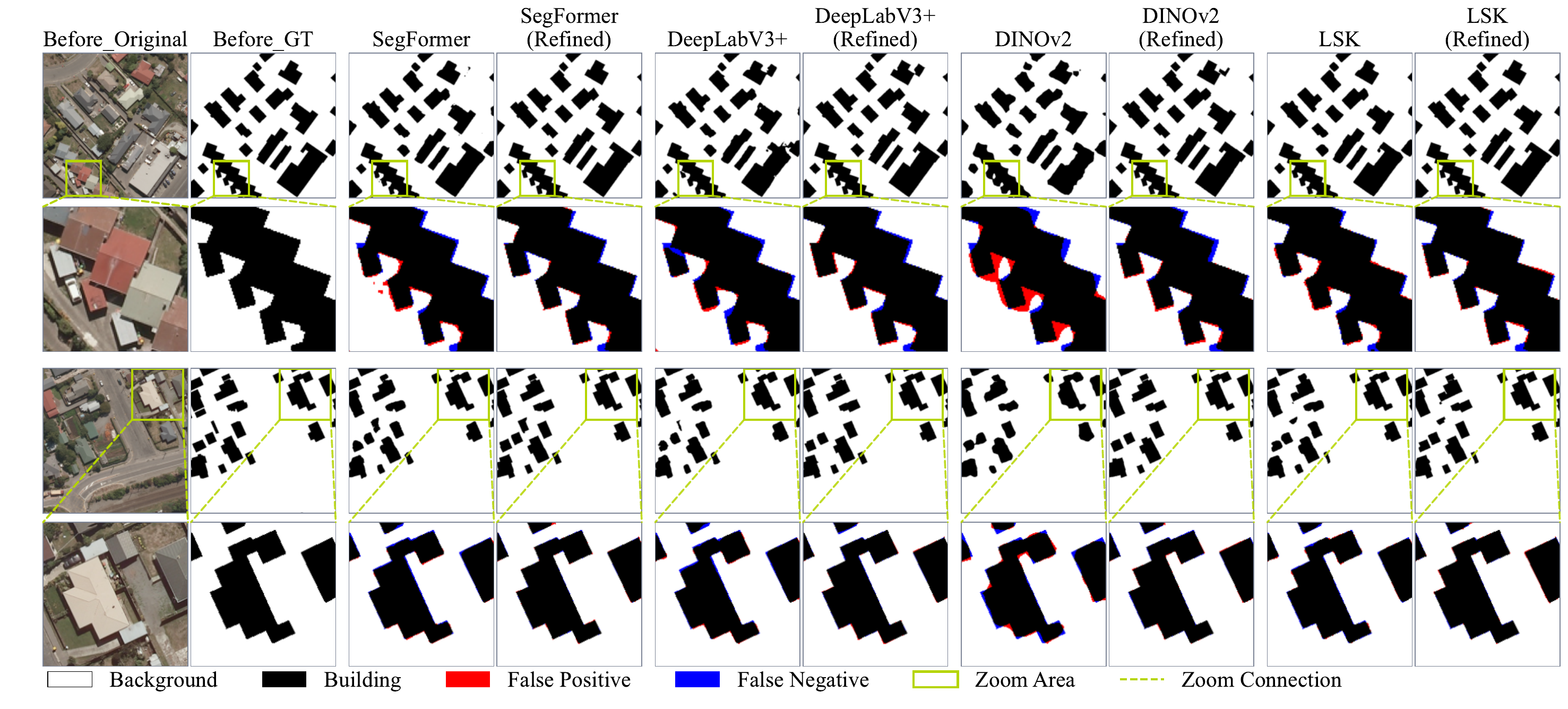}
\caption{Qualitative analysis on the WHU Building post-disaster phase dataset.}
\label{fig:supp_whu_disaster}
\end{figure*}

\begin{figure*}[!t]
\centering
\includegraphics[width=1\textwidth]{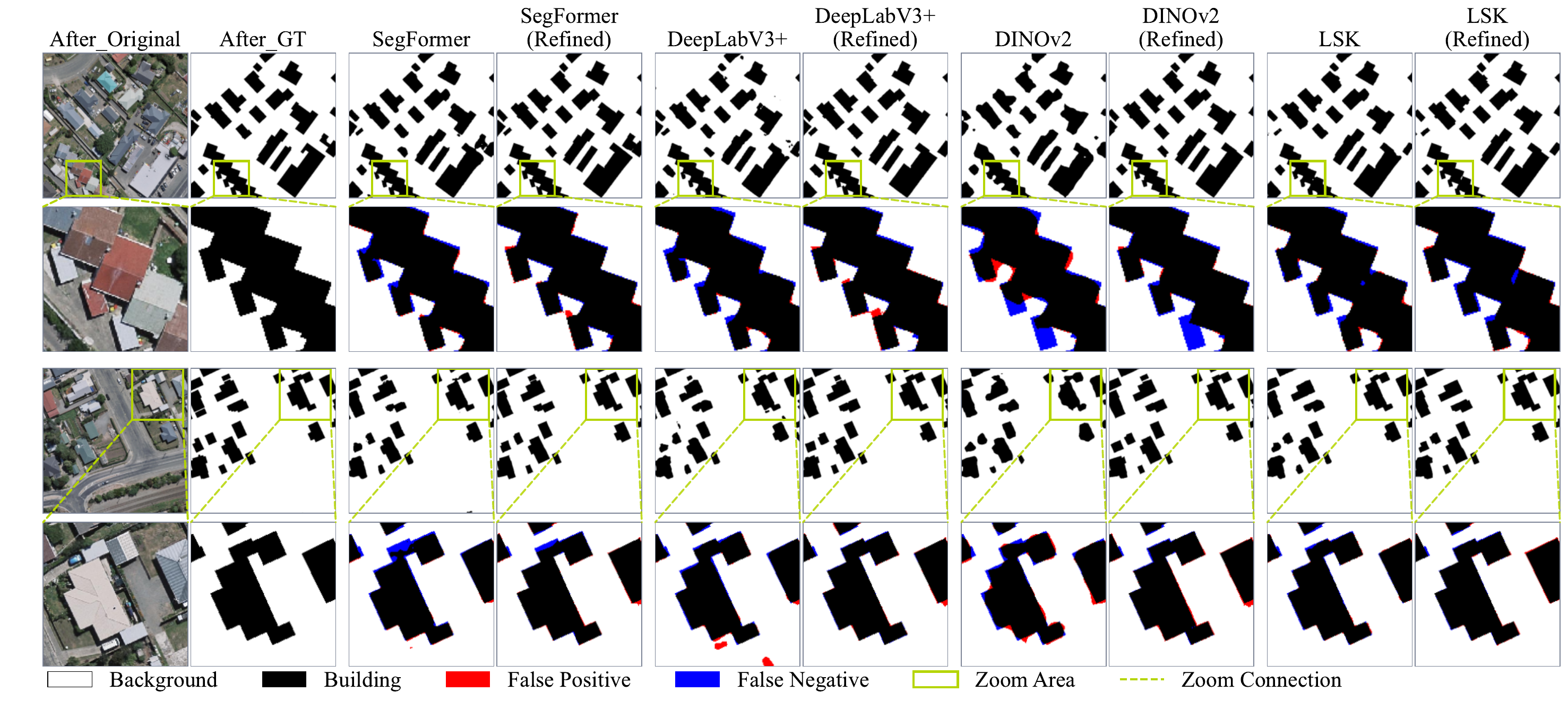}
\caption{Qualitative analysis on the WHU Building post-reconstruction phase dataset.}
\label{fig:supp_whu_reconstruct}
\end{figure*}

Figs.~\ref{fig:supp_whu_disaster} and \ref{fig:supp_whu_reconstruct} demonstrate the qualitative optimization results of IDGBR on post-disaster and post-reconstruction phase images from the WHU Building dataset, respectively. It can be observed that the original discriminative models generally exhibit significant errors near building boundaries, including local false responses, irregular boundary contours, and insufficient geometric structural integrity, particularly in areas with complex roof shapes or blurred boundaries. After introducing IDGBR, all models show significantly reduced errors at boundaries, with smoother and more coherent building edges, and well-preserved detailed morphology, demonstrating the method's clear advantages in boundary modeling. Furthermore, the cross-temporal transfer experiments in Fig.~\ref{fig:supp_whu_reconstruct} verify that models optimized with IDGBR maintain high boundary accuracy during the reconstruction phase, significantly suppressing segmentation errors caused by cross-temporal differences while maintaining stable boundary morphology. This indicates that IDGBR possesses strong temporal stability and cross-scene adaptation capabilities.

\subsection{Qualitative Results on Multi-Class Segmentation Tasks}

To further verify the generalization of our method in complex multi-category scenarios, we visualize the segmentation results on the Potsdam and Vaihingen datasets.

\begin{figure*}[!t]
\centering
\includegraphics[width=1\textwidth]{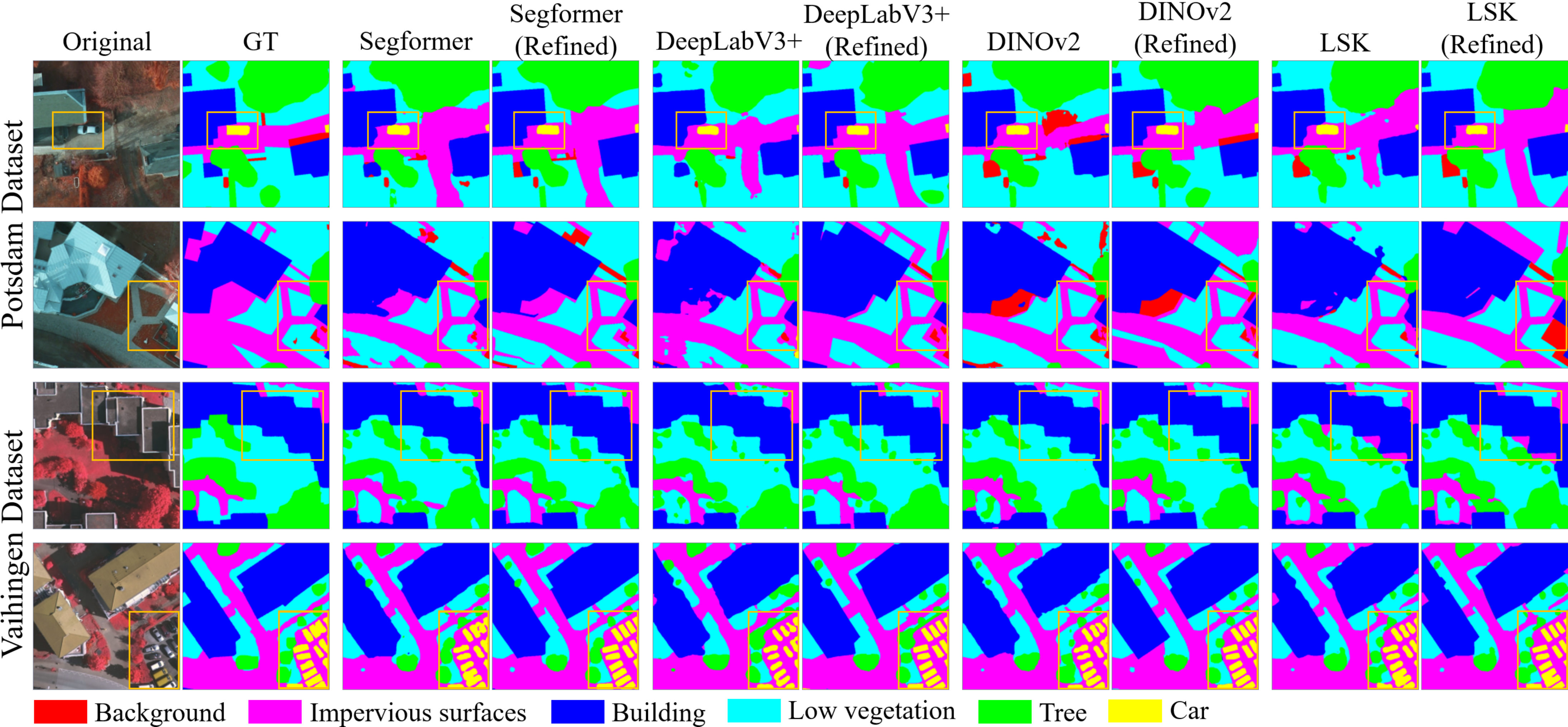}
\caption{Qualitative results of different methods on multi-class segmentation datasets (Potsdam and Vaihingen).}
\label{fig:supp_multiclass}
\end{figure*}

In both Potsdam and Vaihingen datasets, as evidenced in Fig.~\ref{fig:supp_multiclass}, the optimized segmentation results consistently restore spatial structures of discretely distributed small patches and irregular segments (such as vehicle boundaries and vegetation fragments). Additionally, for objects with strong geometric structures (such as buildings and road edges), the consistency between spatial topological relationships and actual ground object distribution is significantly improved after optimization.

\subsection{Inference Efficiency Analysis}
\begin{table*}[t]
\centering
\caption{Performance comparison of cubic sampling strategy on binary datasets with different foreground ratios.}
\label{tab:cubic_binary_analysis}

\setlength{\dashlinedash}{2pt}
\setlength{\dashlinegap}{2pt}
\setlength{\aboverulesep}{0pt}
\setlength{\belowrulesep}{0pt}
\renewcommand{\arraystretch}{1.3}
\setlength{\tabcolsep}{8pt}

\begin{tabular}{l:c c c c c c c}
\toprule
Dataset & Cubic Strategy & IoU & F1-score & OA & Kappa & WFm (3px B) & Foreground Ratio (\%) \\
\midrule
\multirow{2}{*}{CHN6-CUG} & $\checkmark$ & 62.96 & 77.27 & 97.44 & 75.91 & 37.16 & \multirow{2}{*}{7.59} \\
& $\times$ & 64.54 & 78.45 & 97.50 & 77.12 & 37.64 & \\
\midrule
\multirow{2}{*}{FGFD} & $\checkmark$ & 87.52 & 93.35 & 92.32 & 84.27 & 49.24 & \multirow{2}{*}{59.34} \\
& $\times$ & 87.66 & 93.42 & 92.39 & 84.40 & 49.78 & \\
\midrule
\multirow{2}{*}{WHU (Post-disaster Phase)} & $\checkmark$ & 94.76 & 97.31 & 99.19 & 96.83 & 76.46 & \multirow{2}{*}{15.00} \\
& $\times$ & 94.72 & 97.29 & 99.18 & 96.81 & 75.91 & \\
\bottomrule
\end{tabular}
\end{table*}

\begin{table}[htbp]
\centering
\caption{Comparison of GFLOPs and parameters with Conditional Diffusion Model. \label{tab:complexity_comparison}}
\setlength{\dashlinedash}{2pt}
\setlength{\dashlinegap}{2pt}
\setlength{\aboverulesep}{0pt} 
\setlength{\belowrulesep}{0pt}
\renewcommand{\arraystretch}{1.3}
\small
\begin{tabular}{l : c c}
\toprule
\textbf{Methods} & \textbf{GFLOPs} & \textbf{Params (M)} \\
\midrule
Control SD$^{\ast}$ (25 steps) & 24576.92 & 1427.51 \\
Ours (25 steps) & 29121.39 & 1521.09 \\
\bottomrule
\end{tabular}
\begin{flushleft}
\footnotesize
\textit{Note:} ``Control SD$^{\ast}$'' specifically refers to the standard Stable Diffusion v1.5 Model equipped with a structural adapter (i.e., ControlNet architecture). Computational cost (GFLOPs) was obtained by tracking actual operator execution using the fvcore library~\cite{fvcore} in full component loading mode (Batch Size=1, Res=512$\times$512, Steps=25).
\end{flushleft}
\end{table}

Table~\ref{tab:complexity_comparison} presents a comparative analysis of the computational complexity of IDGBR against a conditional diffusion baseline. Diffusion-based methods require an iterative denoising process, which inherently leads to substantial computational overhead. Specifically, compared to the standard conditional diffusion baseline, IDGBR introduces a moderate increase of 18.5\% in GFLOPs and 6.6\% in parameters. The justification for this increased computational cost should be evaluated within the broader context of the full remote sensing mapping lifecycle. This workflow spans from image acquisition and automated interpretation to manual verification, graphic editing, and final map publication. Low-quality segmentation boundaries—manifested as blurring, jagged edges, or topological errors—directly increase the downstream burden of manual verification and editing. By significantly improving boundary quality at a modest computational cost, our method replaces a costly and uncertain manual editing stage with a deterministic, one-time computational investment. Consequently, the overall cost of the mapping process is reduced.

\subsection{Applicability Analysis of Cubic Sampling Strategy in Binary Segmentation Tasks}

Experimental results presented in the main text indicate that the cubic sampling strategy resulted in a slight performance degradation on the CHN6-CUG dataset. Given the extremely low proportion of foreground targets in this dataset (approximately 7.59\%), we preliminarily hypothesized that this phenomenon stemmed from the severe imbalance in sample distribution. To investigate this, we introduced the FGFD dataset (with a balanced foreground ratio of 59.34\%) and the WHU dataset (with a similarly low foreground ratio of 15.00\%) for comparative experiments, as detailed in Table~\ref{tab:cubic_binary_analysis}.

The experimental results reveal that there is no linear correlation between the foreground ratio and the efficacy of the cubic sampling strategy. On the WHU dataset, which features a similarly sparse foreground, cubic sampling yielded a slight performance improvement. Conversely, on the foreground-dominated FGFD dataset, the strategy exerted a suppressive effect on performance. This comparison demonstrates that the performance variations of cubic sampling across datasets cannot be fully attributed solely to the proportion of ground object classes; its performance is affected by additional factors that are not yet fully understood. In binary segmentation tasks, it should be regarded as a condition-sensitive optional optimization strategy.

\end{document}